\documentclass[letterpaper, 10 pt, journal, twoside]{IEEEtran}
\usepackage{CJKutf8}
\usepackage{graphicx}
\usepackage{epsfig}
\usepackage{float}
\usepackage{array}
\usepackage{amsmath,amssymb,amsfonts}
\usepackage{bm}
\usepackage{cite}
\usepackage{enumerate}
\usepackage{caption}
\usepackage[noend]{algpseudocode}
\usepackage{booktabs}
\usepackage{tabularx}
\usepackage{multirow}
\usepackage{ragged2e}
\usepackage{color}
\floatname{algorithm}{Algorithm}

\algnewcommand\algorithmicswitch{\textbf{switch}}
\algnewcommand\algorithmiccase{\textbf{case}}
\algnewcommand\algorithmicassert{\texttt{assert}}
\algnewcommand\Assert[1]{\State \algorithmicassert(#1)}%
\algdef{SE}[SWITCH]{Switch}{EndSwitch}[1]{\algorithmicswitch\ #1\ \algorithmicdo}{\algorithmicend\ \algorithmicswitch}%
\algdef{SE}[CASE]{Case}{EndCase}[1]{\algorithmiccase\ #1}{\algorithmicend\ \algorithmiccase}%
\algtext*{EndSwitch}%
\algtext*{EndCase}%

\newcommand{\eref}[1]{Eq.~(\ref{#1})}
\newcommand{\fref}[1]{Fig.~\ref{#1}}
\newcommand{\sref}[1]{Sec.~\ref{#1}}
\newcommand{\Tref}[1]{Table~\ref{#1}}

\newcolumntype{Z}{>{\centering\let\newline\\\arraybackslash\hspace{0pt}}X}
\newcolumntype{P}[1]{>{\RaggedRight\hspace{0pt}}p{#1}}

\begin{document}

\markboth{IEEE Robotics and Automation Letters. Preprint Version. Accepted January, 2022}%
{Shen \MakeLowercase{\textit{et al.}}: TCL: Tightly Coupled Learning Strategy for Hierarchical Place Recognition}

\author{Yanqing Shen$^{1}$, Ruotong Wang$^{1}$, Weiliang Zuo$^{1}$, and Nanning Zheng$^{1}$
\thanks{Manuscript received: 10, 7, 2021; Revised 12, 13, 2021; Accepted 1, 4, 2022.}
\thanks{This paper was recommended for publication by  Associate Editor C. Cadena upon evaluation of the Associate Editor and Reviewers' comments.
This work was supported by the National Natural Science Foundation of China (Grant No. 61773312, 61790562). \textit{(Corresponding author: Nanning Zheng.)}}
\thanks{$^{1}$Y. Shen, R. Wang, and N. Zheng are with the Institute of Artificial Intelligence and Robotics, Xi'an Jiaotong University, Xi'an, Shaanxi 710049, P.R. China;
{\tt\footnotesize qing1159364090, wrt072 @stu.xjtu.edu.cn; weiliang.zuo, nnzheng@mail.xjtu.edu.cn}}
\thanks{Digital Object Identifier (DOI): see top of this page.}}

\title{TCL: Tightly Coupled Learning Strategy for Weakly Supervised Hierarchical Place Recognition}

\maketitle

\begin{abstract}
Visual place recognition (VPR) is a key issue for robotics and autonomous systems. For the trade-off between time and performance, most of methods use the coarse-to-fine hierarchical architecture, which consists of retrieving top-N candidates using global features, and re-ranking top-N with local features. However, since the two types of features are usually processed independently, re-ranking may harm global retrieval, termed \textit{re-ranking confusion}. Moreover, re-ranking is limited by global retrieval. In this paper, we propose a tightly coupled learning (TCL) strategy to train triplet models. Different from original triplet learning (OTL) strategy, it combines global and local descriptors for joint optimization. In addition, a bidirectional search dynamic time warping (BS-DTW) algorithm is also proposed to mine locally spatial information tailored to VPR in re-ranking. The experimental results on public benchmarks show that the models using TCL outperform the models using OTL, and TCL can be used as a general strategy to improve performance for weakly supervised ranking tasks. Further, our lightweight unified model is better than several state-of-the-art methods and has over an order of magnitude of computational efficiency to meet the real-time requirements of robots.
\end{abstract}


\section{Introduction}
\IEEEPARstart{T}{o}
quickly and accurately recognize the location of a given image (i.e. for a robot), a task called visual place recognition (VPR), is essential for full simultaneous localization and mapping (SLAM) systems.
However, VPR still faces key issues: 1) Appearance, illumination and viewpoint of a place may vary greatly over time.
2) Different locations may look similar, called perceptual aliasing \cite{survey2015}.

\begin{figure}[tbp]
  \centering
  \includegraphics[width=0.45\textwidth]{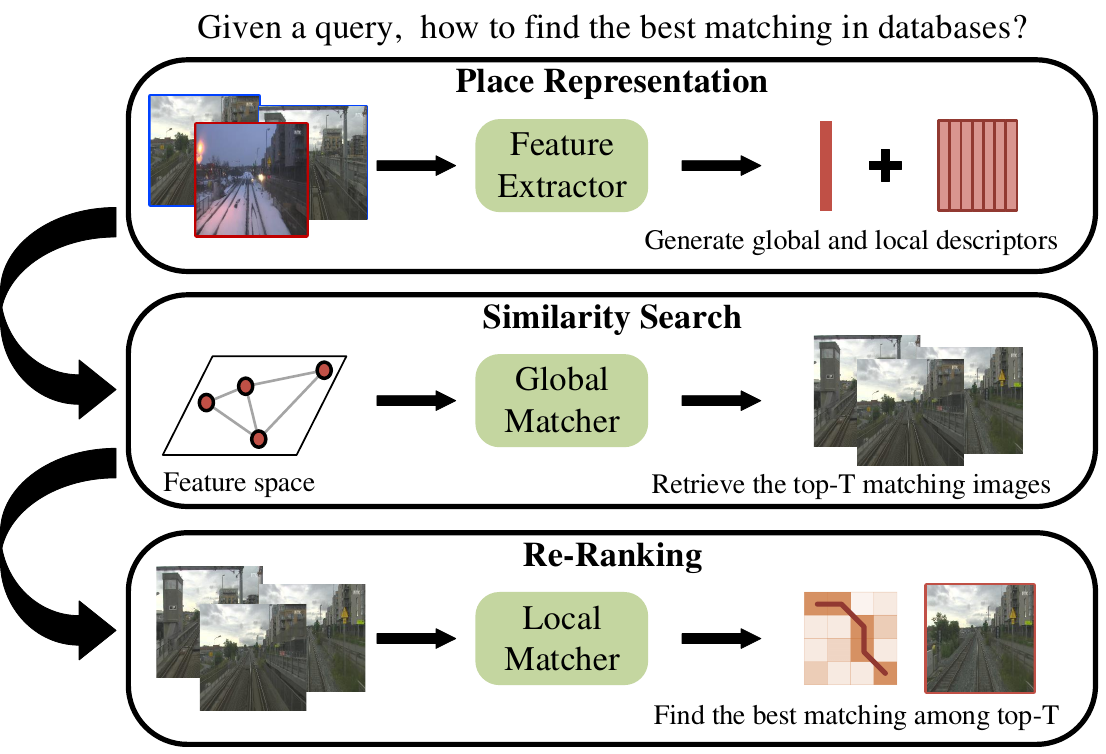}
  \caption{\textbf{Hierarchical retrieval architecture of VPR.} (i) After finetuning a single backbone on VPR task, unified feature extractor provides the global and local descriptors. (ii) Given a query, top-N candidates in database images are retrieved by similarity searching. (iii) The best matching result (R@1) is obtained by local feature-based re-ranking algorithms.}
  \label{fig:arch}
\end{figure}

VPR is typically regarded as an image retrieval task \cite{delf}, which searches for the most similar image to the query in a geo-tagged database.
The common way to represent a single image is to use local or global descriptors. 
Local descriptors \cite{sift,surf,orb,hog} are generated separately for key-points (or regions), and then are used for cross matching \cite{gv,sta}.
Global descriptor-based methods \cite{gist,netvlad} transform an image into a compact vector.
Local descriptors are viewpoint invariant, but encoding and matching require a lot of computational memory and time. Conversely, global descriptors are compact and appearance invariant but suffer from viewpoint changes. 

Considering the complementary advantages of local and global descriptors, 
a widely used architecture is to rank the database by global features, and then re-rank the top candidates using local features, as shown in \fref{fig:arch}. 
Many studies proved that re-ranking could improve the performance \cite{delg,delf,sta,patchnetvlad}.
For example, Patch-NetVLAD \cite{patchnetvlad} derived patch-features from global features for local matching.
However, the fact is that the improvement of re-ranking is limited by global retrieval results, and bad re-ranking may harm ranking results, called \textit{re-ranking confusion}.
This is mainly because the original triplet learning (OTL) strategy separates global and local descriptors, even in a holistic system.

The goal of this paper is to tightly couple global and local descriptors so that they can reinforce each other to improve the global perception capability and enrich the local details, and improve their consistency.
To achieve this goal, we make the following contributions: 1) From the perspective of coupling them in the learning process, we propose a tightly coupled learning (TCL) strategy of triplet networks, and present the re-ranking loss and adapted ranking loss for weakly unsupervised scenario to achieve joint optimization.
2) We deploy TCL on different backbones to train lightweight unified models that combine superior performance with improved computational efficiency, which are also suitable for real-time scenarios.
3) To improve the accuracy of re-ranking, we propose bidirectional search dynamic time warping (BS-DTW) algorithm, with the reasonable path assumptions, to mine more spatial information designed for VPR.

We evaluate the effectiveness of our model and compare with several state-of-the-art methods \cite{netvlad,patchnetvlad,delg,gcl} on well-known datasets (i.e. Mapillary Street Level Sequences (MSLS), Pittsburgh, Nordland and Gardens Point Walking (GPW) \cite{msls,pitts30k,nord1,gardens}).
Ours outperforms speed-focused Patch-NetVLAD and the global feature-based methods across all test sets. Ours performance is better than best-performing Patch-NetVLAD and DELG-local in some datasets while worse in others, but our computational cost is lower (at least 200 times faster). 
We also carry out experiments to prove our model is suitable for real-time scenarios, 
New College \cite{new} and City Centre \cite{city}.
Furthermore, ablation studies are carried out to showcase the effectiveness of TCL and BS-DTW, particularly the potential of TCL to provide more powerful representations that are tailored to the ranking tasks.

\section{Related Work}

\textbf{Global Descriptors.} Global descriptors can provide a compact representation by aggregating local descriptors or directly processing the whole image.
BoW \cite{bow}, Fisher Vectors \cite{fv}, and VLAD \cite{vlad} have been used to aggregate hand-crafted local features through a visual codebook, and then have been incorporated into CNN-based architectures \cite{netvlad}.
Other CNN-based works mainly focused on designing specific pooling layers on top of convolutional trunks \cite{superglue, pool1}.
There have also been some methods using graph-based strategies \cite{add1,add2} to combine semantic and geometric information.

\textbf{Local Descriptors.} The hand-crafted local features such as SIFT \cite{sift}, SURF \cite{surf} and ORB \cite{orb}, and more recent CNN-based features including SuperPoint \cite{superpoint} and D2Net \cite{d2}, have been extensively used in VPR and SLAM systems \cite{city,orbslam, delf,patchnetvlad, seqnet}. 
In addition to local aggregation to obtain global image descriptors, local descriptors are often used to re-rank initial candidates produced by a global approach through cross-matching between image pairs \cite{inloc,gv,sta}. 
Distinct from existing methods, we propose TCL to combine global and local descriptors in learning process to achieve their additional joint optimization.
Most related to our BS-DTW, a non-learning re-ranking algorithm, STA-VPR \cite{sta} proposed an adaptive DTW algorithm to align local features from the spatial domain. However, the assumptions about start and end points are not reasonable enough for VPR.

\section{Proposed Method}

\begin{figure}[tbp]
  \centering
  \includegraphics[width=0.45\textwidth]{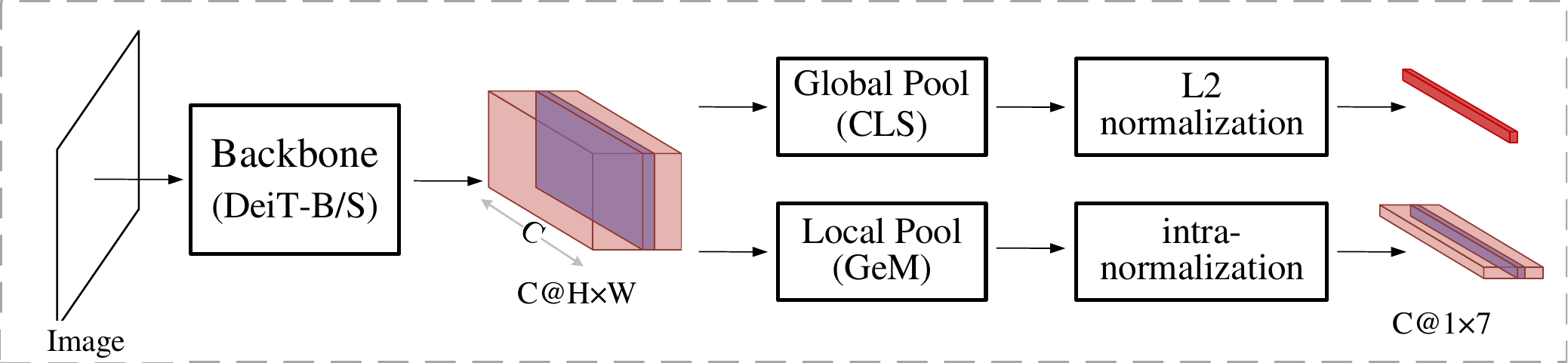}
  \caption{The unified feature extractor that we trained in \fref{fig:TS} consists of a deep network backbone, a global pooling layer and a local pooling layer.}
  \label{fig:feat}
\end{figure}

As illustrated in \fref{fig:feat}, global and local features are extracted from the unified backbone.
TCL strategy makes improvements in tuples mining strategy and loss function definition (\sref{sec:strategy}).
Triplet tuples are selected for the next step of training through the global-local matcher, and adapted ranking loss and additional re-ranking loss are introduced for joint optimization.
In \sref{sec:dtw}, a bidirectional search DTW algorithm is designed for image spatial alignment in re-ranking to provide the local distance.

\subsection{Place Representation}
\label{sec:feat}

\begin{figure*}[htbp]
  \centering
  \includegraphics[width=0.95\textwidth]{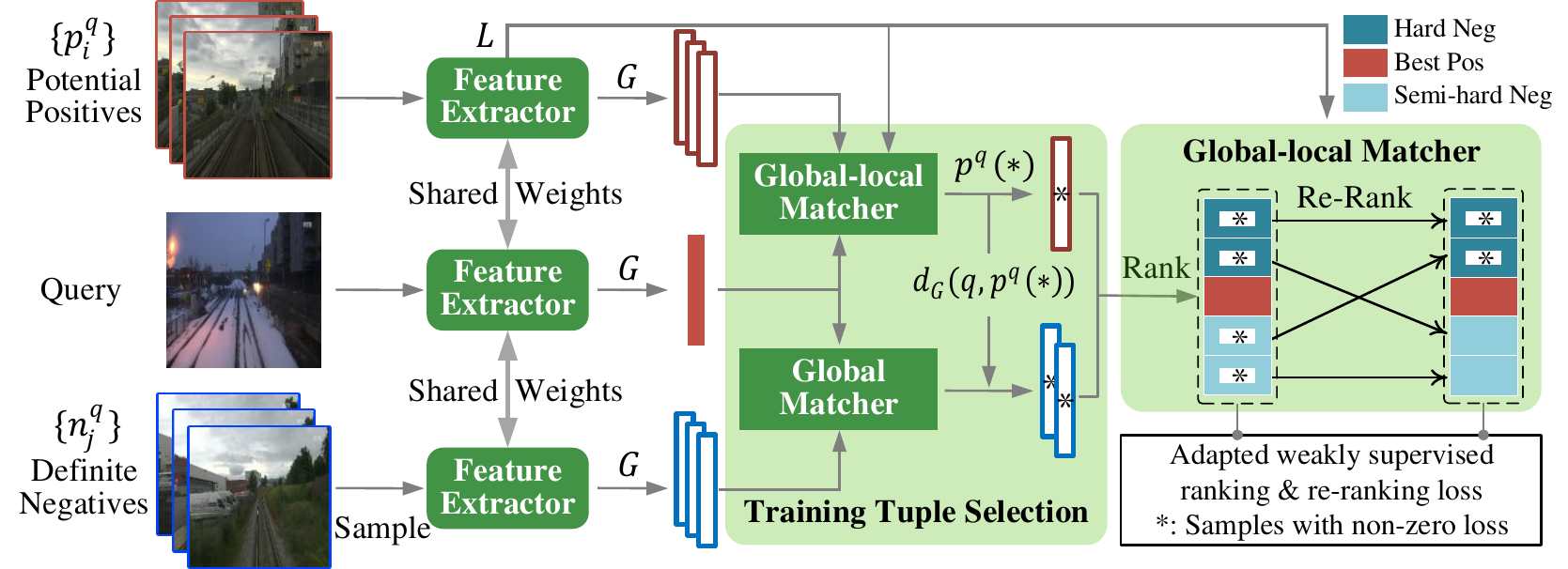}
  \caption{\textbf{Tightly Coupled Learning Strategy of Triplet Networks for Weakly Supervised VPR.} In order to improve performance in the retrieval architecture as \fref{fig:arch}, the TCL strategy couples ranking and re-ranking: selection of training tuples and definition of losses. As shown in the dotted box, \textit{global-local matcher} is a two-stage matching method including ranking and re-ranking instead of only using global ones. In addition, we adapt the triplet ranking loss into our scenario and propose the triplet re-ranking loss. $*$ represents the samples that have non-zero losses.}
  \label{fig:TS}
\end{figure*}

This section introduces the feature extractor and some symbols used in the subsequent description. 
As illustrated in \fref{fig:feat}, global and local features are extracted from a \textbf{unified} model by adding specific pooling layers on a lightweight backbone. The unified model further reduces calculation time and requires no additional or learned local feature extractor.

\textbf{Backbone Network.}
Considering that cascaded self-attention modules in vision transformer (ViT) can capture long-distance feature dependencies, lightweight pretrained ViT models \cite{deit} are used as backbones. They are used to compare with SOTA methods and illustrate the effectiveness and portability of the TCL strategy in ablation studies.

\textbf{Pooling Layer.}
Given an image $I$, the output consists of $M+1$ vectors corresponding to $M$ input patches and a \texttt{[class]} token. Following the spirit of ViT \cite{vit}, we view normalized \texttt{[class]} as a global descriptor, $G(I)$. 

Reshaping the feature map as a $C×H×W$-dimensional (channel by height by width) tensor, we split it vertically into $N$ separate matrixes and apply Generalized Mean (GeM) pooling on each matrix. In this way, an image can be represented as a sequence of $C$-dimensional normalized local features, $\{ L_{1}(I), ..., L_{k}(I),..., L_{N}(I) \}$.

\textbf{Distances.}
\textit{Global distance} is denoted as $d_{G}$ and used for ranking. \textit{Local distance} is denoted as $d_{L}$ and used for re-ranking, with details in \sref{sec:dtw}.

\subsection{Tightly Coupled Learning Strategy}
\label{sec:strategy}

This section introduces the improvements of TCL strategy from two aspects: mining training tuples that are more effective for training speed and model performance, and defining additional loss function.

\textbf{Training Tuples Mining Strategy.}
{
For Google Street View data, each query image $q$ have a set of potential positives $\{p_{i}^{q}\}$ and a set of definite negatives $\{n_{j}^{q}\}$. 
The potential positives mean that there is at least one positive image that's co-visible with $q$, but it's not clear which one it is. To address this problem in the training process, previous mining strategy \cite{netvlad} chose the best matching positive image for each $\left(q,\left\{p_{i}^{q}\right\},\left\{n_{j}^{q}\right\}\right)$ as

\begin{equation}
\label{eq:top}
p_{i*}^{q}=\underset{p_{i}^{q}}{\operatorname{argmin}} \, d_{G}\left(q, p_{i}^{q}\right),
\end{equation}
where $d_{G}$ is the \textbf{global distance} in the current learning phase, $d_{G}(q,p)=\left\|G(q)-G(p)\right\|$.

In order to address the separation of global and local descriptors in learning, we present the \textit{global-local matcher} in our strategy. As shown in \fref{fig:TS}, it chooses the best matching positive image $p^{q}(*)$ in two steps, different from $p_{i*}^{q}$ above.
First, global features are used to select top-$T$ nearest images $\{p_{i*}^{q}\}_T$ from potential positives as \eref{eq:top}. 
Second, local feature-based algorithm is used to re-rank them and obtain top-1 as
\begin{equation}
    p^{q}(*)=\underset{p_{i}^{q}}{\operatorname{argmin}} \, d_{L}\left(q, \{p_{i*}^{q}\}_T \right).
\end{equation}

Subsequently, negatives of hard and semi-hard, $\mathbf{J}_q$, are chosen by \textit{global matcher}, which satisfy 
\begin{equation}
\label{eq:jq}
    d_{G}\left(q, p^{q}(*)\right) + m > d_{G}\left(q, n_{j}^{q}\right), \quad j \in \mathbf{J}_{q},
\end{equation}
where $m$ is a constant parameter giving the margin. Note that we first randomly select a subset of negative samples and then rank them only using global features. 
This is because after measuring the spent time and the gained performance, we find it's not worth re-ranking negative samples.
In other words, the main difference from other strategies lies in the selection of positives.
}

\textbf{Adapted Weakly Supervised Ranking Loss.}
{
In \cite{netvlad}, triplet ranking loss for $\left(q, \left\{p_{i}^{q}\right\}, \left\{n_{j}^{q}\right\}\right)$ is defined as
\begin{equation}
\label{equ:loss}
\sum_{j \in \mathbf{J}_q} h\left(\min _{i} d_{G}\left(q, p_{i}^{q}\right)+m-d_{G}\left(q, n_{j}^{q}\right)\right),
\end{equation}
and we adapt \eref{equ:loss} to our strategy as
\begin{equation}
\label{equ:lossg}
L_{g}=\sum_{j \in J_q} h\left( d_{G}\left(q, p^{q}(*)\right)+m-d_{G}\left(q, n_{j}^{q}\right)\right),
\end{equation}
where $h$ is the hinge loss: $h(x) = \max(x,0)$ and $p^{q}(*)$ is chosen by global-local matcher in \fref{fig:TS}.
}

\textbf{Weakly Supervised Re-Ranking Loss.}
{
To strengthen the coupling between the ranking and re-ranking in learning, an additional loss is defined to constrain optimization space, namely \textit{weakly supervised re-ranking loss}, as shown below:
\begin{equation}
\label{equ:lossdtw}
L_{l}=\sum_{j \in \mathbf{J}_q} h\left(d_{L}\left(q, p^{q}(*)\right)-d_{L}\left(q, n_{j}^{q}\right)\right).
\end{equation}

This loss is proposed from the point of view of recall in testing to improve the accuracy of re-ranking, so only hard negatives for local distance obtain non-zero losses.
It is worth noting that the negative samples in \eref{equ:lossdtw} are still from \eref{eq:jq}, which ensures the consistency with the hierarchical retrieval structure \fref{fig:arch}.

\begin{figure*}[tbp]
  \centering
  \includegraphics[width=\textwidth]{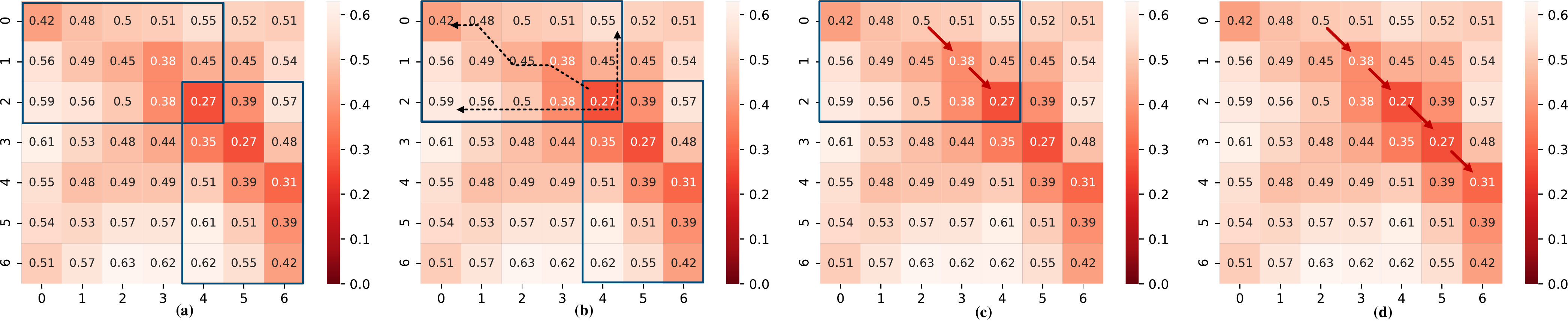}
  \caption{Flow diagram of BS-DTW. The four figures correspond to the steps to find an optimal warping path. In (c), $P_{ul}$ is marked with red arrows.}
  \label{fig:dtw}
\end{figure*}

Finally, the total loss is 
\begin{equation}
    L = w_{g} L_{g} + w_{l} L_{l},
\end{equation}
where $w_g$ and $w_l$ are weights. $L_l$ accelerates the learning process, and makes the parameter space constrained to simultaneously pull out the global and local distances between positives and negatives, thus improving their consistency.
We believe that joint optimization leads to better features that are tailored to the ranking tasks.
}

\subsection{Re-Ranking via BS-DTW}
\label{sec:dtw}

In order to mine spatial alignment information in cross-matching, DTW algorithm has been applied \cite{sta}.
In this section, we present \textit{bidirectional search DTW (BS-DTW)} algorithm, which is designed for local features alignment in image retrieval. It can yield alignment paths and \textbf{local distances $d_L$} between image pairs.

As illustrated in \fref{fig:feat}, each image is represented as a sequence of $N$ local features ($N = 7$ in this paper) in the horizontal dimension, as in STA-VPR \cite{sta}.
As mentioned in \sref{sec:feat}, the distance matrix $\mathbf{D}_{N×N}$ can be established by the distances between pairwise features in two sequences and an optimal path can be obtained:
\begin{equation}
    P=\{ p_{1}, p_{2}, ..., p_{k}, ..., p_{K} \},
\end{equation}
where $N\!<\!K\!<\!2N\!-\!1$ and the path $P$ represents the alignment between two images, $Q$ and $R$.

The path needs to satisfy three characteristics: boundary, continuity and monotonicity. 
The presentation of the real world in different images satisfies the homogeneity, so it is easy to understand that monotonicity and continuity are satisfied in image spatial alignment. 

We focus on analyzing the boundary constraint in spatial alignment. 
Previous DTW algorithms require exact start and end points, e.g., $p_{1}\!=\!(1,1)$ and $p_{K}\!=\!(N,N)$ in \cite{sta}. 
However, through analyzing the matched images from the perspective of co-vision, we find that some segmentation in an image contains something that does not exist in another image.

To alleviate such a limitation, we design a loose boundary constraint for VPR: at least one element in $\{Q1, R1\}$ must have corresponding contents in another image (the same is true for $\{Q7, R7\}$), called \textit{``edge-touching principle''}.
The three constraints can be expressed mathematically as follows:
\begin{itemize}
    \item Boundary: The start point $p_1$ is on the leftmost column or uppermost row, that is, $p_{1}=(1,J_{1})$ or $ (I_{1},1)$, where $ I_{1}, J_{1} \in [1,N]$. Similarly, the end point ${p_K}=(N,J_{N})$ or $(I_{N},N)$, where $I_{N} \in [I_{1},N], J_{N} \in [J_{1},N]$.
    \item Continuity: If $p_{k}=(i,j)$ and $p_{k+1}=(i',j')$, then $i' \leq i+1$ and $j' \leq j+1$. 
    \item Monotonicity: If $p_{k}=(i,j)$ and $p_{k+1}=(i',j')$, then $i \leq i'$ and $j \leq j'$. 
\end{itemize}

Without the need for exact start and end points, BS-DTW more reasonably assumes that the path needs to contain the position with the highest local similarity, theoretically corresponding to the minimum value in $\mathbf{D}_{N×N}$.
The computation is summarized as follows:
\begin{enumerate}
    \item Given $\mathbf{D}_{N×N}$ between two images, find the minimum value and the corresponding index $(i^{*},j^{*})$.
    To reduce the risk of extreme incorrect selection, we add additional judgement that more than two of 8-neighborhood are in the top-13 minimum of $\mathbf{D}_{N×N}$.
    Then two exploration spaces are generated as shown in \fref{fig:dtw}(a) and $P$ can be divided into two sub-paths, $P_{ul}$ for upper-left part and $P_{lr}$ for lower-right part:
$$
    \begin{aligned}
        P_{ul} &= \{ p_{1}, ..., (i^{*},j^{*}) \}, \\
        P_{lr} &= \{(i^{*},j^{*}), ..., p_{K} \}.
    \end{aligned}
$$
    \item Taking $P_{ul}$ as an example, potential start point satisfies $p'_{1}\in  \mathbf{S}_{1}$, where $\mathbf{S}_{1}=\{ (1,j), (i,1) \}$ and $i \in [1,i^{*}], j \in [1,j^{*}]$. For every $p'_{1}$, the optimal path to $(i^{*},j^{*})$ and the normalized distance are obtained by \eref{equ:path} and (\ref{equ:norm}).
    \item Select the point with the shortest normalized distance in $\mathbf{S}_{1}$ as the start point $p_{1}$ in $P$. 
    \item Repeat step 2)-3) for $P_{lr}$ and obtain the end point $p_{K}$, and path $P$ is shown in \fref{fig:dtw}(d).
    \item Compute the normalized distance of $P$ as the \textit{local distance, $d_L$}.
\end{enumerate}

Note that the above distance is calculated by the standard DTW methods \cite{shapedtw}, and the distance $s_{i, j}$ of the optimal path from $(1,1)$ to $(i,j)$ is shown in this formula, referring to the expression of Eq.(3) in \cite{sta}:
\begin{equation}
\begin{tiny}
\label{equ:path}
s_{i,j}=
\begin{cases}
d_{i, j} & i=1, j=1 \\ 
d_{i, j}+s_{i, j-1} & i=1, j \in[2,N] \\ 
d_{i, j}+s_{i-1, j} & i \in[2,N], j=1 \\ 
d_{i, j} + \min \left\{ s_{i-1, j-1}, s_{i-1, j}, s_{i, j-1}\right\} & i, j \in[2,N] 
\end{cases},
\end{tiny}
\end{equation}
where $ d_{i, j}$ is the Euclidean distance between local features, $L_{i}(Q)$ and $L_{j}(R)$.
The normalized local distance between two images is 
\begin{equation}
\label{equ:norm}
    d_{L} = \frac{s_{N,N}}{l_P},
\end{equation}
where $l_P$ is the length of the optimal path.
In our algorithm, the loose boundary constraints make the top-left and bottom-right vertices not mandatory to be included in the path, which is more in line with the practical situations.

\section{Experiments}

In this section we first describe the datasets and evaluation metrics, and details about our implementation.
We then evaluate TCL strategy and BS-DTW via detailed ablation studies, and compare our trained lightweight unified model with state-of-the-art place recognition methods.
Finally, we show the results of our model in real-time SLAM scenarios. 

\subsection{Datasets and Evaluation}
We test models on several benchmark datasets and one synthetic dataset, using their recommended configuration and public partition of validation/test sets.
A summary of datasets is shown in \Tref{tbl:dataset}. Here we introduce the usage of datasets to facilitate an informed assessment of the results.
\textbf{Mapillary Street Level Sequences (MSLS)} \cite{msls} contains over 1.6 million images recorded in urban and suburban areas over 7 years. We evaluate models on the validation set.
\textbf{Pittsburgh (Pitts30k)} \cite{pitts30k} contains 30k database images and 24k queries, which are geographically divided into train/validation/test sets \cite{netvlad}. We evaluate models on the test set.
\textbf{Nordland} \cite{nordland} are divided into train and test set \cite{nord1}, and down-sampled images ($224×224$) of test set in summer (reference) and winter (query) are used\footnote{Same as previous works \cite{patchnetvlad,nord1}, we remove all black tunnels and times when the train is stopped.}.
\textbf{Synth-Nord} is synthesized by cropping the right 30\% of databases and the left 10\% of queries as \cite{sta}.
\textbf{Gardens Point Walking (GPW)} \cite{gardens} captures two traverses during the day and night. The day-left (query) and night-right (reference) traverses are used.
\textbf{City Centre} \cite{city} and \textbf{New College} \cite{new} are used for testing real-time SLAM in \sref{sec:slam}.

\begin{table}[tbp]
  \caption{Summary of the datasets. No. refers to the number of images. ++ indicates severe changes and + indicates medium changes.}
  \label{tbl:dataset}
  \begin{center}
  \begin{tabularx}{8.5cm}{crccc}
    \toprule
      \multirow{2}{*}{Dataset} & \multirow{2}{*}{No. (R/Q)} & \multirow{2}{*}{Description} & \multicolumn{2}{c}{Variation} \\
      \cline{4-5} &&&App.& View. \\
      \hline
    \midrule
      MSLS-train\cite{msls} & 900k/500k & \multirow{2}{1.7cm}{urban,suburban, \\ long-time} &++ & ++\\
      MSLS-val & 19k/11k &  &++ &++ \\
      \hline
      Pitts30k-train\cite{pitts30k} & 10k/7,416 & \multirow{2}{1.7cm}{urban, \\ pitch variation} & ++ & ++ \\
      Pitts30k-test & 10k/6,816 &  & ++ & ++ \\
      \hline
      Nordland \cite{nord1} & 3,450/3,450 & \multirow{2}{1.7cm}{train journey, \\ seasonal} & ++ & none \\
      synth-Nord & 3,450/3,450 & & ++ & ++ \\
      \hline
      GPW \cite{gardens} & 200/200 & \multirow{1}{1.7cm}{campus} & ++ & + \\
    \bottomrule
  \end{tabularx}
\end{center}
\end{table}

\textbf{Evaluation Metrics.}
We evaluate the recognition performance mainly based on Recall@N, whereby query is regarded to be correctly localized if at least one of the top N retrieved database images is within the ground truth tolerance. In addition, we evaluate the performance of SLAM by precision-recall scores. 
Follow the recommended evaluation of datasets, ground truth matches are obtained by given binary matrix from authors for New College \cite{new} and City Centre \cite{city}, 25m translation error for Pitts30k \cite{netvlad,patchnetvlad}, 25m translation and 40$^ \circ$ orientation error for MSLS \cite{msls}, and 5 closet corresponding frames (i.e. $\pm 2$) for two-path-form datasets, i.e. Nordland \cite{nord1} and GPW \cite{gardens}.

\subsection{Implementation}
As \cite{patchnetvlad}, we train models on Pitts30k for urban imagery (Pitts30k and GPW), and MSLS for other conditions.

Following the standard data augmentation methods, each image is resized to $(256,256)$ with random horizontal flipping, and then randomly cropped to $(224,224)$.
We use two transformer-based models as backbones to build lightweight unified models. The pre-trained weights on ImageNet \cite{imagenet} are from the public implementation1 of DeiT \cite{deit} built on the timm library\footnote{https://github.com/rwightman/pytorch-image-models}.
Our models are finetuned with triplet network using single parameter configuration with the margin $m=0.1$ and Adam optimizer using a learning rate of $0.000005$, a weight decay of $0.0001$ and a batch size of 2.
All experiments are run on an NVIDIA GeForce RTX 2080 Ti, except for training on MSLS with two cards.
The finetuned model with the best recall@5 on the val set is used to test datasets, and the top-10 and top-100 images are respectively re-ranked in our experiments, termed -R10 and -R100.

For \sref{sec:strategy}, potential positives mean the database images within 10 meters, and definite negatives mean images further away than 25 meters. 
$T=5$ in global-local matcher and 10 hardest negatives are chosen from 1000 randomly sampled negatives. Considering that global distances and local distances are close, $w_{g}=w_{l}=0.5$.

\subsection{Ablation Studies}

\begin{table*}[tbp]
    \begin{center}
    \caption{Results of different strategies on Pitts30k, Nordland, and GPW. All models use our BS-DTW to re-rank top-100 unless marked with $\dagger$, which indicates the use of adaptive DTW of STA-VPR \cite{sta}. All models are finetuned on Pitts30k for Pitts30k and GPW, marked with $*$, and MSLS for Nordland marked with $**$. Training time refers to the time taken for the best R@5 on Pitts30k val. set.}
    \label{tbl:ablation}
    \centering
     \begin{tabularx}{17.1cm}{c|c|c|c|c|ccc||ccc||ccc}
    \toprule
       \multirow{2}{*}{Backbone} & \multirow{2}{*}{\#Dim} & \multirow{2}{*}{\#Params} &\multirow{2}{*}{Strategies} & Training & \multicolumn{3}{c||}{\textbf{Pittsburgh30k test$^*$}} & \multicolumn{3}{c||}{\textbf{Gardens Point Walking$^*$}} &
       \multicolumn{3}{c}{\textbf{Nordland test$^{**}$}} \\
        \cline { 6 - 14 } &&&&Time& R@1 & R@5 & R@10 & R@1 & R@5 & R@10 & R @1 & R@5 & R@10\\
    \midrule
       \multirow{6}{*} {DeiT-B} & \multirow{6}{*} {768}  & \multirow{6}{*} {86M}  
       &NL&/&35.92&59.82&70.32&43.00&75.00&88.00&36.99&41.45&44.00\\
        &&& OTL&13h53m&77.88&93.48&96.31&54.50&79.00&90.50&40.14&47.59&49.99\\
        &&& OTL-R$^{\dagger}$ & 13h53m & 79.85 & 93.51 & 96.32  & 57.00 & 80.50 &91.00& 41.75 & 48.52 & 50.11 \\        &&& OTL-R&13h53m&81.55&93.80&96.45&58.50&81.50&94.00&42.61&48.90&50.28\\
        &&&TCL&2h42m&82.51&94.83&97.13&62.00&86.50&94.00&43.32&52.68&59.20\\
        &&&\textbf{TCL-R}&2h42m& \textbf{86.67}&\textbf{95.01}&\textbf{97.15}&\textbf{70.00}&\textbf{88.00}&\textbf{95.00}&\textbf{45.61}&\textbf{54.70}&\textbf{60.26}\\
         \hline
       \multirow{6}{*} {DeiT-S}&\multirow{6}{*} {384}  & \multirow{6}{*} {22M}  
      &NL&/&39.22&63.41&73.74&44.50&75.50&87.00&36.17&40.00&42.49\\
        &&& OTL&8h38m&78.06 &93.09&95.83&55.00&79.00&91.50&39.16&44.61&48.64\\
        &&& OTL-R$^{\dagger}$ & 8h38m & 79.37 & 93.21 & 96.02  & 57.00 & 81.50 &91.00&41.39&45.94&48.84 \\ &&& OTL-R&8h38m&81.57&93.82&96.23&59.00&84.50&92.50&42.38&46.26&49.54\\
        &&&TCL&1h54m&82.40&94.72&97.00&63.00&93.00&95.00&42.75 & 50.52 & 53.73 \\
        &&&\textbf{TCL-R}&1h54m&\textbf{90.51}&\textbf{95.94}&\textbf{97.45}&\textbf{80.50}&\textbf{95.00}&\textbf{97.00} &\textbf{43.45}&\textbf{51.72}&\textbf{54.26}\\
    \bottomrule
  \end{tabularx}
\end{center}
\end{table*}

{\textbf{Learning strategy.}
From three perspectives of whether to finetune model, how to finetune, and whether to add re-ranking as backend in testing, we design ablated variants of \textbf{weakly supervised} strategies on two base architectures:

\begin{itemize}
    \item \textbf{NL}: No learning strategy.
    \item \textbf{OTL}: Original triplet learning strategy \cite{netvlad}.
    \item \textbf{OTL-R}: OTL with Re-Ranking.
    \item \textbf{TCL}: Tightly coupled learning strategy (ours).
    \item \textbf{TCL-R}: TCL with Re-Ranking.
\end{itemize}

\begin{figure}[tbp]
  \centering
  \includegraphics[width=0.45\textwidth]{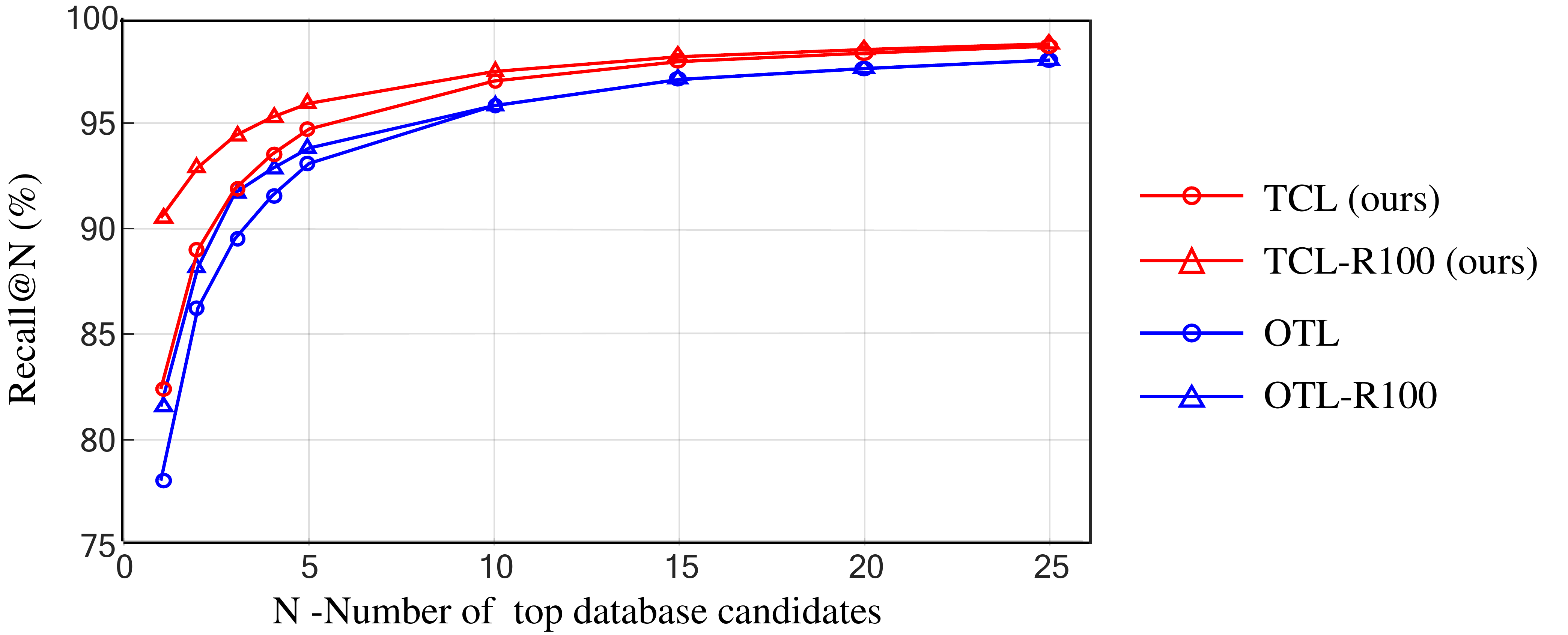}
  \caption{Comparisons of DeiT-S on the Pitts30k-test dataset. We show the Recall@N performance of TCL in red and OTL in blue.}
  \label{fig:RN}
\end{figure}

\Tref{tbl:ablation} shows the comparison results with above variations in terms of training and testing performance. 
TCL reduces the training time by a relative $75\%$ over OTL. That is because our learning strategy focuses on selecting more accurate positives during weakly supervised training, and then gradients will be in a more accurate direction to accelerate the convergence with less epochs.
Note that the training of TCL-R is same as TCL.

The following numeric analysis on performance are based on R@1 – generally applicable to R@5 and R@10. The percentage of Recall is plotted for different values of N in \fref{fig:RN}.
DeiT-S trained on Pitts30k performs a little better than DeiT-B, while DeiT-B trained on MSLS is better. That's because the model representation ability should match the size of training dataset to achieve the model generalization ability.
The performance of ranking and re-ranking increases significantly on Pitts30k and GPW, while on Nordland is not obvious. This shows that our strategy has more advantages in datasets with viewpoint changes.

Through analyzing the data, we can draw the following conclusions.
First, TCL provides more powerful global image representation than OTL for ranking task, thus raising the upper limit of final performance.
Second, the gap between TCL to TCL-R and OTL to OTL-R shows that TCL also improves the perception ability of local features and alleviates re-ranking confusion.
Last, since the above two points are both valid on different backbones, TCL can be used as a general strategy for weakly supervised ranking tasks.
}

\begin{figure}[tbp]
  \centering
  \includegraphics[width=0.45\textwidth]{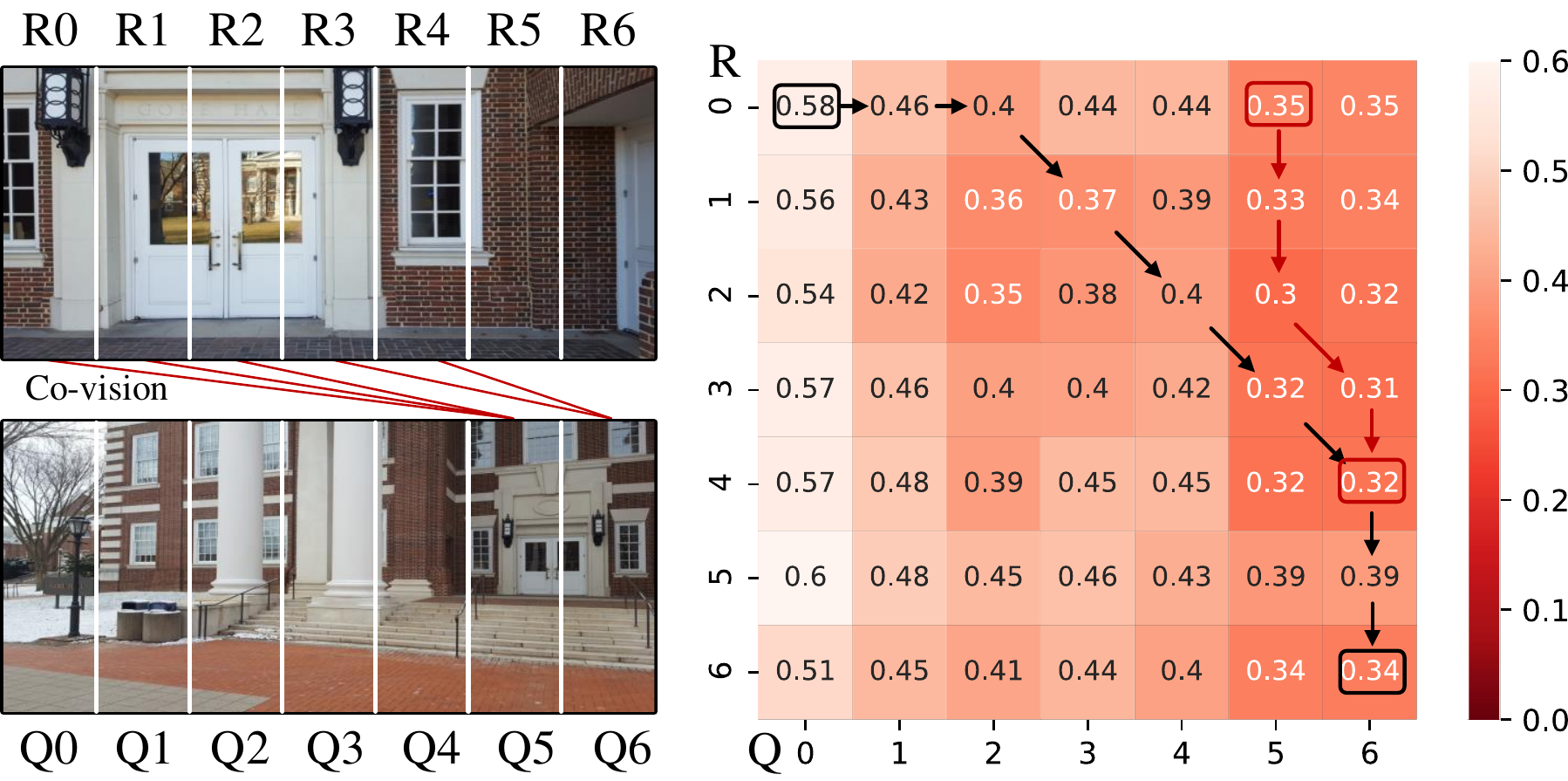}
  \caption{Qualitative example of warping paths. The red box and arrows indicate ours and the black ones indicate the path by adaptive DTW of STA-VPR \cite{sta}.}
  \label{fig:dtw1}
\end{figure}

{\textbf{Re-Ranking Method.}
We conduct experiments to compare BS-DTW with adaptive DTW of STA-VPR \cite{sta}. Extended on the same backbone (DeiT-B/S), the two methods are used as a stand-alone spatial alignment backend, differing only in the way of re-ranking. The results of using adaptive DTW of STA-VPR are shown as OTL-R$^{\dagger}$ in \Tref{tbl:ablation}. 
The results show that our algorithm results in better \textit{re-ranking} performance (average relative increase of 80$\%$), especially on Pitts30k and GPW which have viewpoint changes. This indicates that BS-DTW yields a more accurate alignment path when there is a large viewpoint change.
It's worth noting again that the limited improvement of re-ranking is due to the confusion problem mentioned above and the limitation of global retrieval accuracy.

In \fref{fig:dtw1}, a co-vision image pair with large viewpoint and scale changes is shown on the left. $Q1 - Q4$ contains contents that does not exist in $R$, violating the assumption that $p_1$ and $p_K$ are fixed in paths.
The optimal path of two methods on the right highlights the rationality and flexibility of BS-DTW at the start and end points.

}

\subsection{Comparisons with State-of-the-Art Methods}

We compare with several retrieval-based localization solutions on image-to-image task: methods with re-ranking (DELG \cite{delg} and Patch-NetVLAD \cite{patchnetvlad}) and methods without re-ranking (NetVLAD \cite{netvlad}, GCL \cite{gcl} and DELG-global).

For results not presented in the original paper of baselines, we test their publicly available models (e.g., GCL (ResNet152-GeM-PCA) and Patch-NetVLAD (speed-focused/best-performing, denoted as -s/-p)), and follow their \textbf{optimal testing configuration}, such as PatchNetVLAD use models of different training sets to test datasets (mapillary/pittsburgh\_WPCA) and ours (DeiT-B-MSLS and DeiT-S-Pitts).
Re-Ranking is performed on top-100 candidates.

\begin{figure}[tbp]
  \centering
  \includegraphics[width=0.45\textwidth]{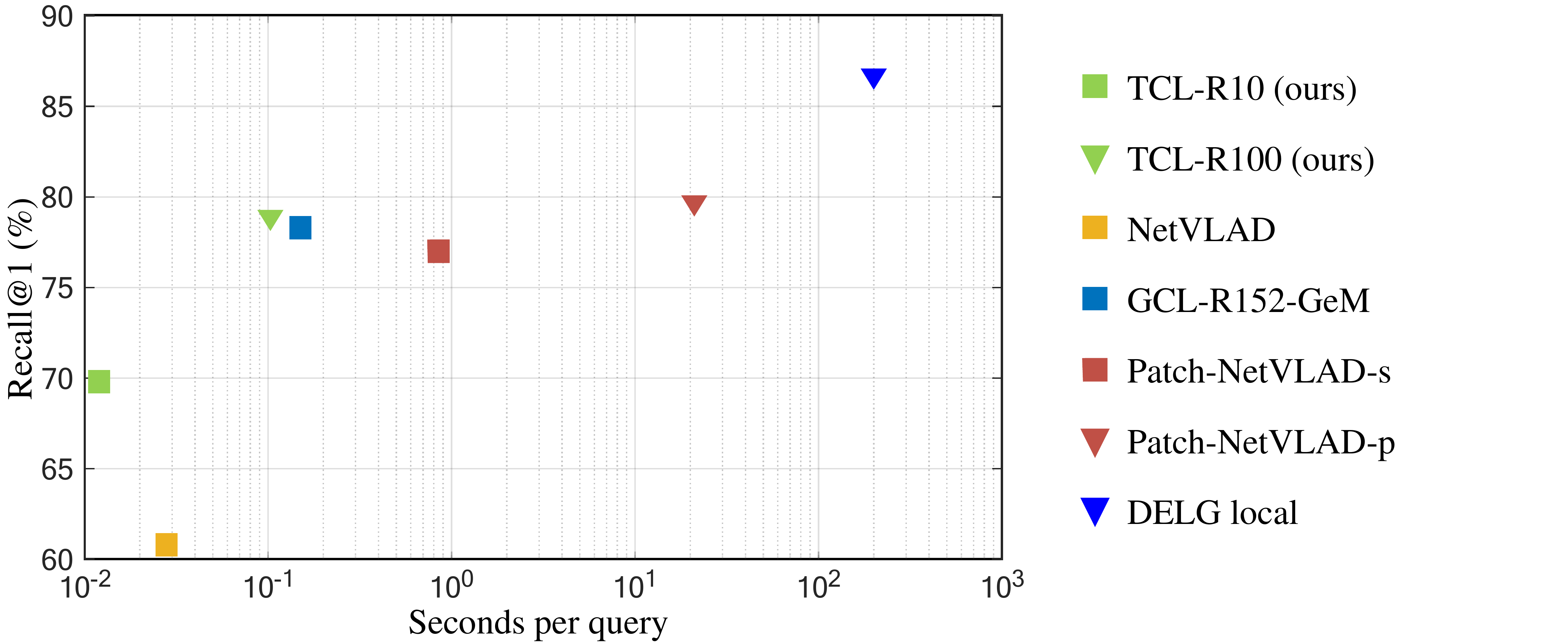}
  \caption{The accumulated time required to process one query image is shown on the x-axis (log), with the R@1 shown on the y-axis, for MSLS-val.}
  \label{fig:RNtime}
\end{figure}

\begin{table*}[tbp]
\setlength\tabcolsep{5.1pt}
\begin{scriptsize}
    \begin{center}
    \caption{Comparison of the quantitative results. $\ddagger$ indicates models trained on the Pitts250k. The average accumulated time required for each query on Pitts30k-test dataset is obtained on an NVIDIA GeForce RTX 2080 Ti.}
    \label{tbl:sota}
    \centering
     \begin{tabularx}{17.5cm}{c|c|c|c|ccc||ccc||ccc||ccc}
    \toprule
       \multirow{2}{*}{Method} & \multirow{2}{*}{Dim} & Img & Time & \multicolumn{3}{c||}{\textbf{Pittsburgh30k test}} & \multicolumn{3}{c||}{\textbf{synth-Nord}}  & \multicolumn{3}{c||}{\textbf{Gardens Point Walking}}& \multicolumn{3}{c}{\textbf{MSLS val}}\\ 
    \cline { 5 - 16 } &&Size&(s)& R@1 & R@5 & R@10 & R@1 & R@5 & R@10 & R @1 & R@5 & R@10 & R @1 & R@5 & R@10\\
    \midrule
    NetVLAD$^\ddagger$ \cite{netvlad}  &32768 & original & 0.028 &  81.9 & 91.2 & 93.7 & 12.4 & 26.4&33.9 & 38.0& 74.5& 83.5 & 60.8 & 74.3 & 79.5\\ 
    GCL \cite{gcl} & 2048 & (640,480) & 0.15 & 75.6 & 91.5 & 92.6 & 18.7 & 28.1 &35.8& 55.5&85.5&93.0 & 78.3&85.0&86.9 \\ 
    DELG global \cite{delg} &128 &(256,256)& - & 80.4 & 90.0 & 93.4 &36.1&49.3&53.2&69.5&92.0&94.0&72.4 &81.9&85.7  \\
    Patch-NetVLAD-s \cite{patchnetvlad} & 512 & (640,480)& 0.85 &87.1& 94.0 & 95.6& 38.2 &48.2 &51.2& 71.5& 93.5 & 96.5 & 77.0 &84.2&86.9 \\
    Patch-NetVLAD-p \cite{patchnetvlad} & 4096 & (640,480) & 21.04 & 88.7& 94.5 & 95.9 & 39.2 & 52.3 &57.1 &78.0 & 94.0&\textbf{97.0} &79.5&86.2&87.7\\ 
    DELG local \cite{delg} &128 &(256,256)& 200 & 90.0 & 95.7 & 97.0 &\textbf{45.9}&\textbf{53.7}&\textbf{59.2}&79.0&\textbf{95.0}&\textbf{97.0}& \textbf{86.5} &\textbf{90.3}&\textbf{91.9} \\
    \hline\hline
    \textbf{Ours}-R10& 384 &(224,224)& \textbf{0.012} & 90.1 & 95.4 & 97.0 & 38.3 & 48.7 & 51.8& 67.5& 84.0& 91.0&69.8&76.3&82.2 \\
    \textbf{Ours}-R100& 384&(224,224)& 0.103& \textbf{90.5}& \textbf{95.9}& \textbf{97.5}&42.3&\textbf{53.7}&56.5& \textbf{80.5} &\textbf{95.0} &\textbf{97.0} &78.7 &82.5 &85.3 \\
    \bottomrule
  \end{tabularx}
\end{center}
\end{scriptsize}
\end{table*}

\begin{figure*}[tbp]
  \centering
  \includegraphics[width=0.9\textwidth]{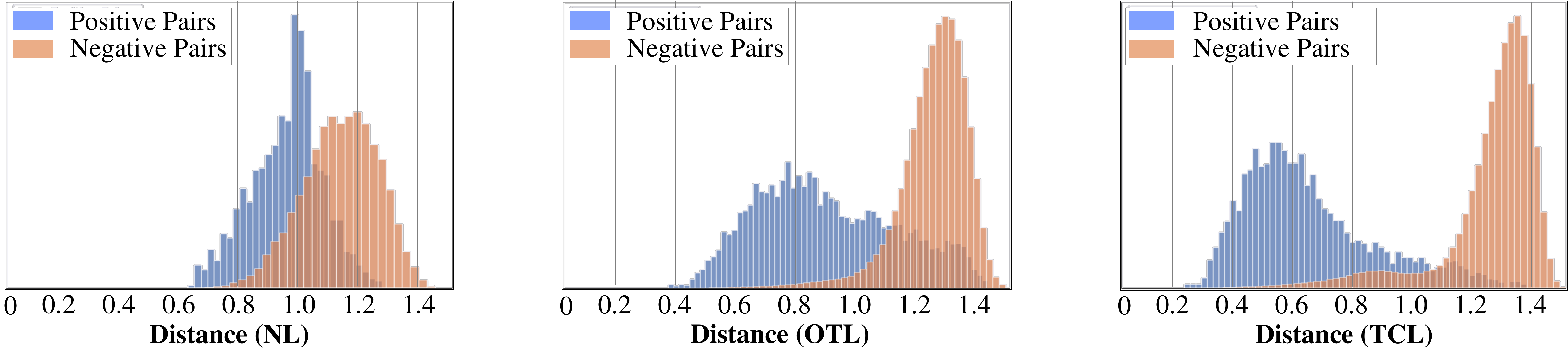}
  \caption{Histograms of the distances between positive and negative pairs. The overlap in TCL is much smaller than NL and OTL.}
  \label{fig:histogram}
\end{figure*}

\Tref{tbl:sota} contains quantitative comparisons with the baseline methods from the perspective of performance and computational efficiency.
The accumulated time includes both feature encoding time and feature matching time\footnote{Matching time means the time taken to find the final R@1 in the database and is proportional to the size of database, which is 10,000 in our testing.}.

Our method (ours-R100) outperforms all global feature methods, NetVLAD, GCL, DELG-global and Patch-NetVLAD-s on datasets on average by $23.2\%$, $14.5\%$, $6.9\%$ and $3.5\%$ (absolute improvement for R@1). Even ours-R10 yielded better results than these global methods, especially on Pitts30k and GPW.

Compared with systems which utilize local feature re-ranking, our method has competitive performance.
Our method yields considerably higher performance on the urban imagery with distinctive structures, i.e. Pitts30k and GPW.
Patch-NetVLAD and DELG achieves better performance on MSLS because of their more complex network structures. But they also comes at a high computational cost on memory and time, and are not suitable for real-time applications.
As shown in \Tref{tbl:sota}, our computation is about 8, 200 and 1900 times faster than Patch-NetVLAD-s, Patch-NetVLAD-p and DELG-local.
\fref{fig:RNtime} gives an intuitive result that our algorithm achieves a good balance of accuracy and time.
 
Let's analyze why our strategy works. First, more accurate tuple mining strategy and fine optimization constraints in weakly supervised training will help the network learn more accurate information. Second, TCL strategy effectively separates positive and negative samples in the feature space. As shown in \fref{fig:histogram}, compared with NL and OTL, TCL reduces the overlap in positive and negative histogram distribution. 

\begin{figure*}[tbp]
  \centering
  \includegraphics[width=0.9\textwidth]{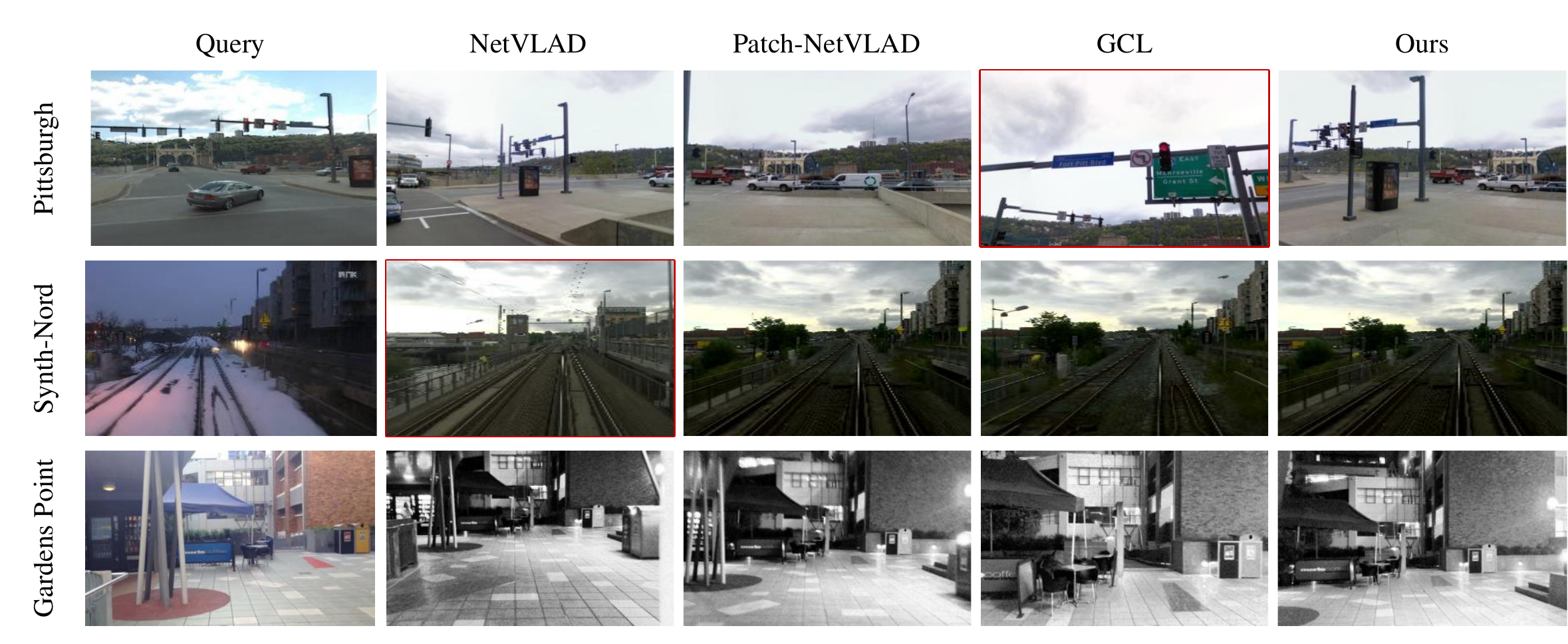}
  \caption{Qualitative Results. For each query, the R@1 results by our model and other SOTA methods are presented. In these examples, ours and Patch-NetVLAD successfully retrieve the matching reference image, while both NetVLAD and GCL produce some incorrect matches marked with \textcolor{red}{red} borders.}
  \label{fig:match}
\end{figure*}

In \fref{fig:match}, we show some matching examples retrieved by our method and SOTA methods.

\begin{table}[htbp]
\begin{footnotesize}
    \begin{center}
    \caption{Performance on the New College and City Centre datasets. We compare several SOTA algorithms with relatively high real-time performance.}
    \label{tbl:slam}
    \centering
     \begin{tabularx}{8.7cm}{c|cc||cc}
    \toprule
        \multirow{2}{*}{Method} & \multicolumn{2}{c||}{New College} & \multicolumn{2}{c}{City Centre}\\
       \cline{2-5}  & Recall(\%) & Time(ms) & Recall(\%) & Time(ms) \\
    \midrule
        NetVLAD & 75.1 & 17.2 & 66.7 & 16.9\\
        GCL & 82.0 & 33.6 & 53.0 & 33.2 \\
        PatchNetVLAD-s & \textbf{91.2} & 272.2 & {84.0} & 289.0 \\
        ours & 89.2 & \textbf{3.6} & \textbf{85.6} & \textbf{3.5} \\
    \bottomrule
  \end{tabularx}
\end{center}
\end{footnotesize}
\end{table}

\subsection{Applications on Robots}
\label{sec:slam}

In order to prove that our model can be applied to robotics, whether it is loop closure detection or re-localization, we carry out experiments on two challenging datasets, i.e. New College \cite{new} and City Centre \cite{city}. Note that only the left camera image stream is used for evaluation.

\begin{figure}[tbp]
  \centering
  \includegraphics[width=0.45\textwidth]{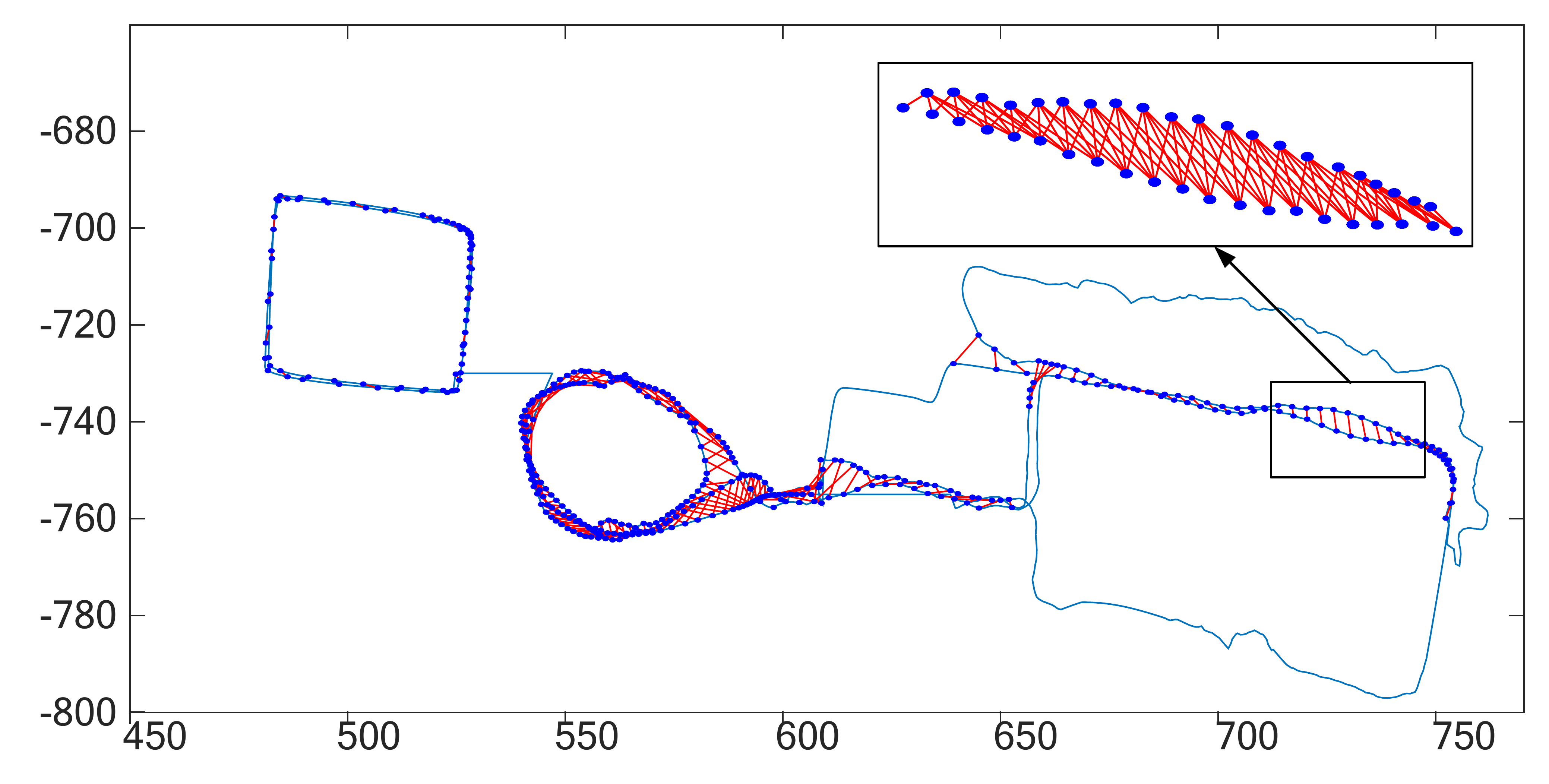}
  \caption{Loop closure detection results (red lines) of New College.}
  \label{fig:slam}
\end{figure}

We select the \textbf{top10} retrieval results for each query for PR analysis.
The maximum recalls at 100\% precision and average execution time of our method (DeiT-S-R10) are shown in \Tref{tbl:slam}. It illustrates that our model can provide accurate and real-time results with appropriate threshold. 
In \fref{fig:slam}, we show the detected matches.

More results can be viewed in the attached video.

\subsection{Discussion}
We found that no single model performed well on all datasets, as was the case with GCL and Patch-NetVLAD, due to the lack of training on one dataset covering the variation of viewpoint, environment and appearance.
Moreover, significant perceptual aliasing and extreme viewpoint variations in testing make image-to-image VPR difficult, even for humans.

We consider that training generalization on different datasets is acceptable with some prior knowledge of the test dataset.
For extremely challenging cases, we believe that adding semantic or sequential information might be helpful.

\section{Conclusions}
In this paper, we propose a tightly coupled learning strategy for weakly supervised hierarchical place recognition.
In order to solve the separation between global and local features in hierarchical architecture, we tightly couple the two-style features in learning - (i) selecting the training tuples and (ii) defining the loss function - to obtain the features tailored to the ranking tasks beyond the place recognition task.
Further, we also propose a bidirectional search DTW algorithm for spatial alignment in re-ranking.
Detailed ablation studies validate the portability and effectiveness of the proposed learning strategy and spatial alignment algorithm.
Further experimental results show that our lightweight unified model trained using the proposed strategy outperforms several state-of-the-art methods on standard benchmarks and is much more suitable for the actual SLAM scenarios. 

\bibliographystyle{IEEEtran}
\bibliography{IEEEabrv,ref}



\end{document}


\title{Supplementary information of ``Realization of fast all-microwave CZ gates with a tunable coupler'' }

\author{Shaowei Li}
\affiliation{Department of Modern Physics, University of Science and Technology of China, Hefei 230026, China}
\affiliation{Shanghai Branch, CAS Center for Excellence in Quantum Information and Quantum Physics, University of Science and Technology of China, Shanghai 201315, China}
\affiliation{Shanghai Research Center for Quantum Sciences, Shanghai 201315, China}

\author{Daojin Fan}
\affiliation{Department of Modern Physics, University of Science and Technology of China, Hefei 230026, China}
\affiliation{Shanghai Branch, CAS Center for Excellence in Quantum Information and Quantum Physics, University of Science and Technology of China, Shanghai 201315, China}
\affiliation{Shanghai Research Center for Quantum Sciences, Shanghai 201315, China}

\author{Ming Gong}
\affiliation{Department of Modern Physics, University of Science and Technology of China, Hefei 230026, China}
\affiliation{Shanghai Branch, CAS Center for Excellence in Quantum Information and Quantum Physics, University of Science and Technology of China, Shanghai 201315, China}
\affiliation{Shanghai Research Center for Quantum Sciences, Shanghai 201315, China}

\author{Yangsen Ye}
\affiliation{Department of Modern Physics, University of Science and Technology of China, Hefei 230026, China}
\affiliation{Shanghai Branch, CAS Center for Excellence in Quantum Information and Quantum Physics, University of Science and Technology of China, Shanghai 201315, China}
\affiliation{Shanghai Research Center for Quantum Sciences, Shanghai 201315, China}

\author{Xiawei Chen}
\affiliation{Department of Modern Physics, University of Science and Technology of China, Hefei 230026, China}
\affiliation{Shanghai Branch, CAS Center for Excellence in Quantum Information and Quantum Physics, University of Science and Technology of China, Shanghai 201315, China}
\affiliation{Shanghai Research Center for Quantum Sciences, Shanghai 201315, China}

\author{Yulin Wu}
\affiliation{Department of Modern Physics, University of Science and Technology of China, Hefei 230026, China}
\affiliation{Shanghai Branch, CAS Center for Excellence in Quantum Information and Quantum Physics, University of Science and Technology of China, Shanghai 201315, China}
\affiliation{Shanghai Research Center for Quantum Sciences, Shanghai 201315, China}

\author{Huijie Guan}
\affiliation{Department of Modern Physics, University of Science and Technology of China, Hefei 230026, China}
\affiliation{Shanghai Branch, CAS Center for Excellence in Quantum Information and Quantum Physics, University of Science and Technology of China, Shanghai 201315, China}
\affiliation{Shanghai Research Center for Quantum Sciences, Shanghai 201315, China}

\author{Hui Deng}
\affiliation{Department of Modern Physics, University of Science and Technology of China, Hefei 230026, China}
\affiliation{Shanghai Branch, CAS Center for Excellence in Quantum Information and Quantum Physics, University of Science and Technology of China, Shanghai 201315, China}
\affiliation{Shanghai Research Center for Quantum Sciences, Shanghai 201315, China}

\author{Hao Rong}
\affiliation{Department of Modern Physics, University of Science and Technology of China, Hefei 230026, China}
\affiliation{Shanghai Branch, CAS Center for Excellence in Quantum Information and Quantum Physics, University of Science and Technology of China, Shanghai 201315, China}
\affiliation{Shanghai Research Center for Quantum Sciences, Shanghai 201315, China}

\author{He-Liang Huang}
\affiliation{Department of Modern Physics, University of Science and Technology of China, Hefei 230026, China}
\affiliation{Shanghai Branch, CAS Center for Excellence in Quantum Information and Quantum Physics, University of Science and Technology of China, Shanghai 201315, China}
\affiliation{Shanghai Research Center for Quantum Sciences, Shanghai 201315, China}

\author{Chen Zha}
\affiliation{Department of Modern Physics, University of Science and Technology of China, Hefei 230026, China}
\affiliation{Shanghai Branch, CAS Center for Excellence in Quantum Information and Quantum Physics, University of Science and Technology of China, Shanghai 201315, China}
\affiliation{Shanghai Research Center for Quantum Sciences, Shanghai 201315, China}

\author{Kai Yan}
\affiliation{Department of Modern Physics, University of Science and Technology of China, Hefei 230026, China}
\affiliation{Shanghai Branch, CAS Center for Excellence in Quantum Information and Quantum Physics, University of Science and Technology of China, Shanghai 201315, China}
\affiliation{Shanghai Research Center for Quantum Sciences, Shanghai 201315, China}

\author{Shaojun Guo}
\affiliation{Department of Modern Physics, University of Science and Technology of China, Hefei 230026, China}
\affiliation{Shanghai Branch, CAS Center for Excellence in Quantum Information and Quantum Physics, University of Science and Technology of China, Shanghai 201315, China}
\affiliation{Shanghai Research Center for Quantum Sciences, Shanghai 201315, China}

\author{Haoran Qian}
\affiliation{Department of Modern Physics, University of Science and Technology of China, Hefei 230026, China}
\affiliation{Shanghai Branch, CAS Center for Excellence in Quantum Information and Quantum Physics, University of Science and Technology of China, Shanghai 201315, China}
\affiliation{Shanghai Research Center for Quantum Sciences, Shanghai 201315, China}

\author{Haibin Zhang}
\affiliation{Department of Modern Physics, University of Science and Technology of China, Hefei 230026, China}
\affiliation{Shanghai Branch, CAS Center for Excellence in Quantum Information and Quantum Physics, University of Science and Technology of China, Shanghai 201315, China}
\affiliation{Shanghai Research Center for Quantum Sciences, Shanghai 201315, China}

\author{Fusheng Chen}
\affiliation{Department of Modern Physics, University of Science and Technology of China, Hefei 230026, China}
\affiliation{Shanghai Branch, CAS Center for Excellence in Quantum Information and Quantum Physics, University of Science and Technology of China, Shanghai 201315, China}
\affiliation{Shanghai Research Center for Quantum Sciences, Shanghai 201315, China}

\author{Qingling Zhu}
\affiliation{Department of Modern Physics, University of Science and Technology of China, Hefei 230026, China}
\affiliation{Shanghai Branch, CAS Center for Excellence in Quantum Information and Quantum Physics, University of Science and Technology of China, Shanghai 201315, China}
\affiliation{Shanghai Research Center for Quantum Sciences, Shanghai 201315, China}

\author{Youwei Zhao}
\affiliation{Department of Modern Physics, University of Science and Technology of China, Hefei 230026, China}
\affiliation{Shanghai Branch, CAS Center for Excellence in Quantum Information and Quantum Physics, University of Science and Technology of China, Shanghai 201315, China}
\affiliation{Shanghai Research Center for Quantum Sciences, Shanghai 201315, China}

\author{Shiyu Wang}
\affiliation{Department of Modern Physics, University of Science and Technology of China, Hefei 230026, China}
\affiliation{Shanghai Branch, CAS Center for Excellence in Quantum Information and Quantum Physics, University of Science and Technology of China, Shanghai 201315, China}
\affiliation{Shanghai Research Center for Quantum Sciences, Shanghai 201315, China}

\author{Chong Ying}
\affiliation{Department of Modern Physics, University of Science and Technology of China, Hefei 230026, China}
\affiliation{Shanghai Branch, CAS Center for Excellence in Quantum Information and Quantum Physics, University of Science and Technology of China, Shanghai 201315, China}
\affiliation{Shanghai Research Center for Quantum Sciences, Shanghai 201315, China}

\author{Sirui Cao}
\affiliation{Department of Modern Physics, University of Science and Technology of China, Hefei 230026, China}
\affiliation{Shanghai Branch, CAS Center for Excellence in Quantum Information and Quantum Physics, University of Science and Technology of China, Shanghai 201315, China}
\affiliation{Shanghai Research Center for Quantum Sciences, Shanghai 201315, China}

\author{Jiale Yu}
\affiliation{Department of Modern Physics, University of Science and Technology of China, Hefei 230026, China}
\affiliation{Shanghai Branch, CAS Center for Excellence in Quantum Information and Quantum Physics, University of Science and Technology of China, Shanghai 201315, China}
\affiliation{Shanghai Research Center for Quantum Sciences, Shanghai 201315, China}

\author{Futian Liang}
\affiliation{Department of Modern Physics, University of Science and Technology of China, Hefei 230026, China}
\affiliation{Shanghai Branch, CAS Center for Excellence in Quantum Information and Quantum Physics, University of Science and Technology of China, Shanghai 201315, China}
\affiliation{Shanghai Research Center for Quantum Sciences, Shanghai 201315, China}

\author{Yu Xu}
\affiliation{Department of Modern Physics, University of Science and Technology of China, Hefei 230026, China}
\affiliation{Shanghai Branch, CAS Center for Excellence in Quantum Information and Quantum Physics, University of Science and Technology of China, Shanghai 201315, China}
\affiliation{Shanghai Research Center for Quantum Sciences, Shanghai 201315, China}

\author{Jin Lin}
\affiliation{Department of Modern Physics, University of Science and Technology of China, Hefei 230026, China}
\affiliation{Shanghai Branch, CAS Center for Excellence in Quantum Information and Quantum Physics, University of Science and Technology of China, Shanghai 201315, China}
\affiliation{Shanghai Research Center for Quantum Sciences, Shanghai 201315, China}

\author{Cheng Guo}
\affiliation{Department of Modern Physics, University of Science and Technology of China, Hefei 230026, China}
\affiliation{Shanghai Branch, CAS Center for Excellence in Quantum Information and Quantum Physics, University of Science and Technology of China, Shanghai 201315, China}
\affiliation{Shanghai Research Center for Quantum Sciences, Shanghai 201315, China}

\author{Lihua Sun}
\affiliation{Department of Modern Physics, University of Science and Technology of China, Hefei 230026, China}
\affiliation{Shanghai Branch, CAS Center for Excellence in Quantum Information and Quantum Physics, University of Science and Technology of China, Shanghai 201315, China}
\affiliation{Shanghai Research Center for Quantum Sciences, Shanghai 201315, China}

\author{Na Li}
\affiliation{Department of Modern Physics, University of Science and Technology of China, Hefei 230026, China}
\affiliation{Shanghai Branch, CAS Center for Excellence in Quantum Information and Quantum Physics, University of Science and Technology of China, Shanghai 201315, China}
\affiliation{Shanghai Research Center for Quantum Sciences, Shanghai 201315, China}

\author{Lianchen Han}
\affiliation{Department of Modern Physics, University of Science and Technology of China, Hefei 230026, China}
\affiliation{Shanghai Branch, CAS Center for Excellence in Quantum Information and Quantum Physics, University of Science and Technology of China, Shanghai 201315, China}
\affiliation{Shanghai Research Center for Quantum Sciences, Shanghai 201315, China}

\author{Cheng-Zhi Peng}
\affiliation{Department of Modern Physics, University of Science and Technology of China, Hefei 230026, China}
\affiliation{Shanghai Branch, CAS Center for Excellence in Quantum Information and Quantum Physics, University of Science and Technology of China, Shanghai 201315, China}
\affiliation{Shanghai Research Center for Quantum Sciences, Shanghai 201315, China}

\author{Xiaobo Zhu}
\email{xbzhu16@ustc.edu.cn}
\affiliation{Department of Modern Physics, University of Science and Technology of China, Hefei 230026, China}
\affiliation{Shanghai Branch, CAS Center for Excellence in Quantum Information and Quantum Physics, University of Science and Technology of China, Shanghai 201315, China}
\affiliation{Shanghai Research Center for Quantum Sciences, Shanghai 201315, China}

\author{Jian-Wei Pan}
\affiliation{Department of Modern Physics, University of Science and Technology of China, Hefei 230026, China}
\affiliation{Shanghai Branch, CAS Center for Excellence in Quantum Information and Quantum Physics, University of Science and Technology of China, Shanghai 201315, China}
\affiliation{Shanghai Research Center for Quantum Sciences, Shanghai 201315, China}

\date{\today}

\pacs{03.65.Ud, 03.67.Mn, 42.50.Dv, 42.50.Xa}

\maketitle
\beginsupplement

\section{Numerical simulation of CZ gate}

\begin{figure*}[hb]
    \centering
    \includegraphics[width=0.8\textwidth]{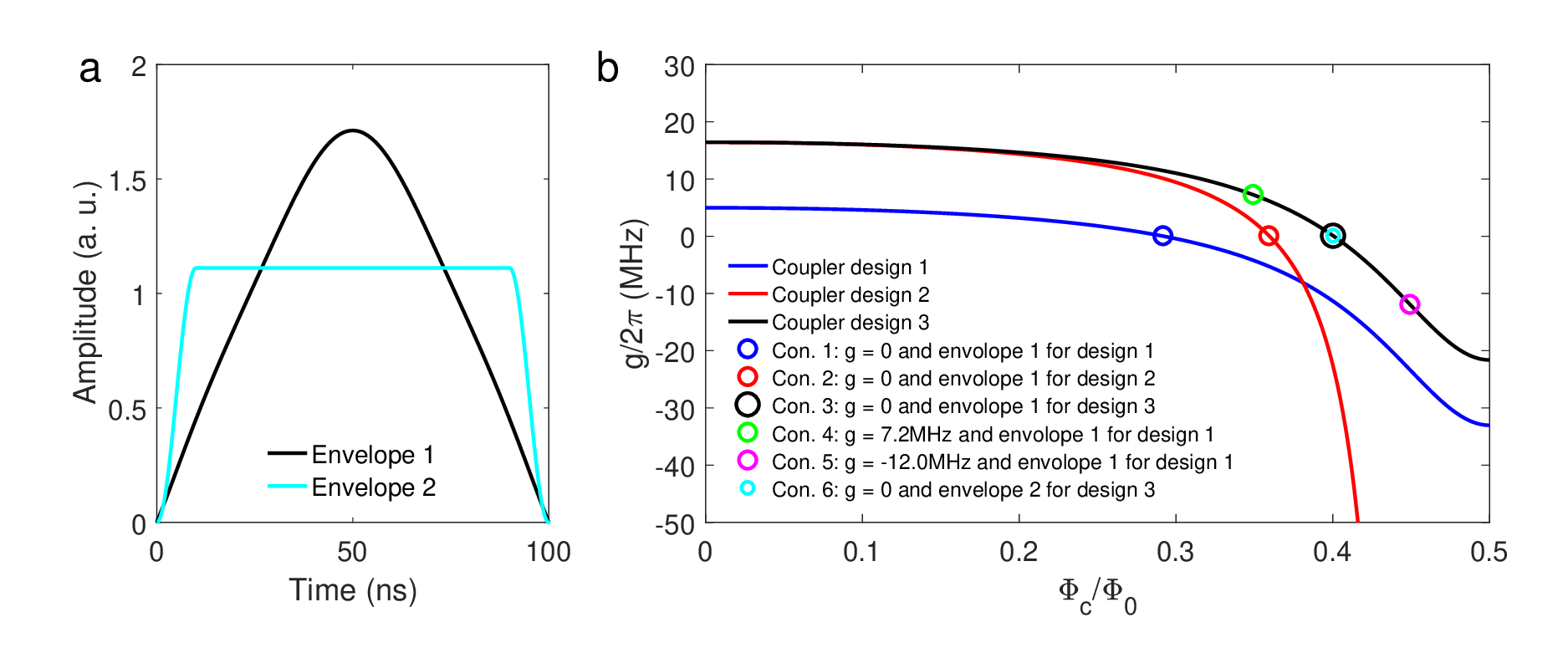}
    \caption{Different coupler designs, idle points and initial envelopes for condition 1 (con. 1) to condition 6. }
    \label{Figure_S1}
\end{figure*}

\begin{figure*}[hb]
    \centering
    \includegraphics[width=0.8\textwidth]{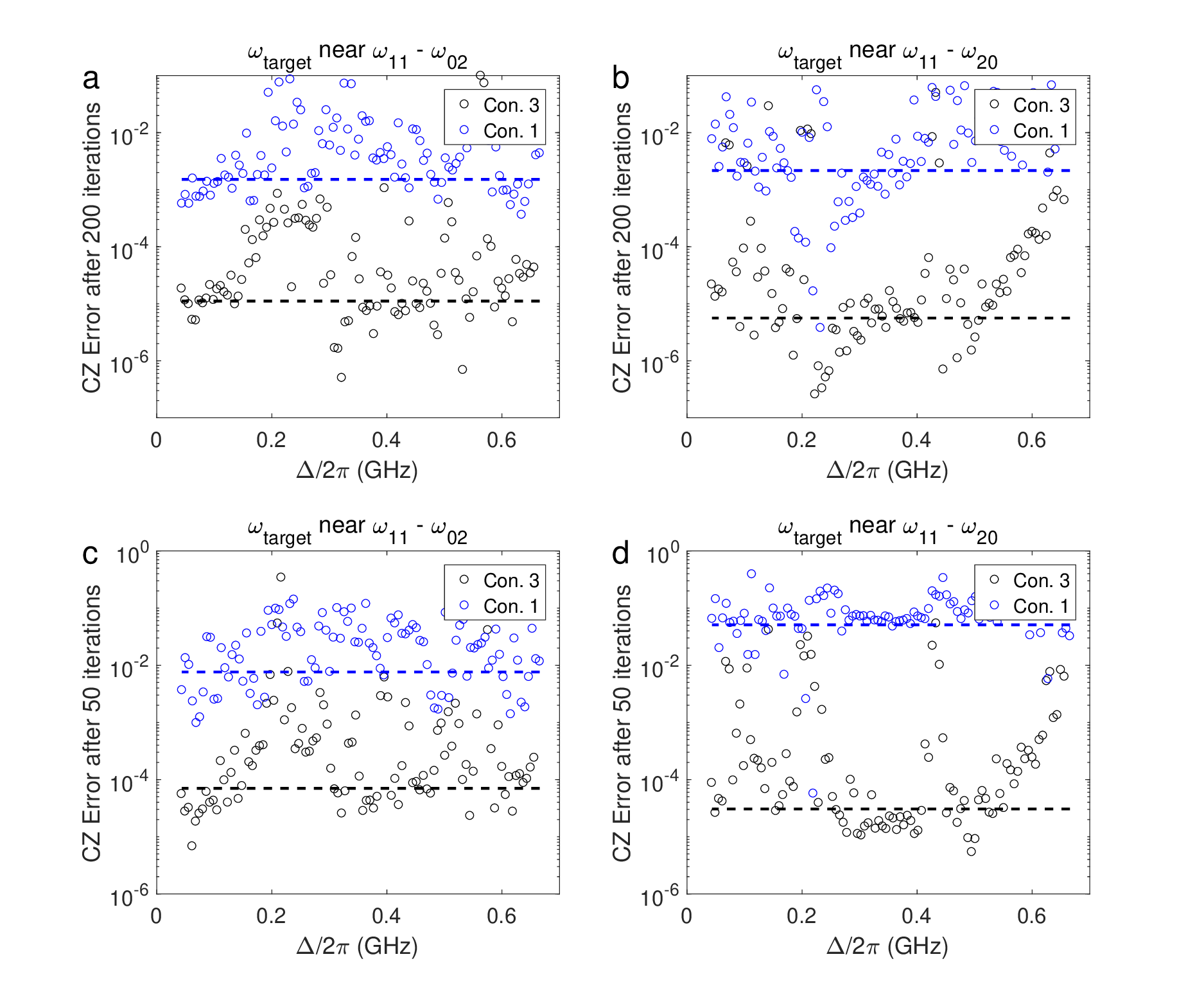}
    \caption{Results of Con. 3 and Con. 1 of CZ optimization. (a) CZ error under the con. 1 and con. 3 for selecting $\omega_{target}$ near $\omega_{11} - \omega_{02}$. The circles are CZ errors under con. 3 and con. 1 after 200 iteration of Nelder Mead algorithm for different $\Delta$. For some of $\Delta$, the unnecessary resonance interactions will increase the gate error, see Fig. 2 (d) in main text. We selected the best 50 from the results of all 100 different $\Delta$ to calculate the average error rate, as shown by the dash line. (b) CZ error under the con. 1 and con. 3 for selecting $\omega_{target}$ near $\omega_{11} - \omega_{20}$ after 200 iteration. (c) CZ error under the con. 1 and con. 3 for selecting $\omega_{target}$ near $\omega_{11} - \omega_{02}$ after 50 iteration. (d) CZ error under the con. 1 and con. 3 for selecting $\omega_{target}$ near $\omega_{11} - \omega_{20}$ after 50 iteration.}
    \label{Figure_S2}
\end{figure*}

\begin{figure*}[hb]
    \centering
    \includegraphics[width=0.8\textwidth]{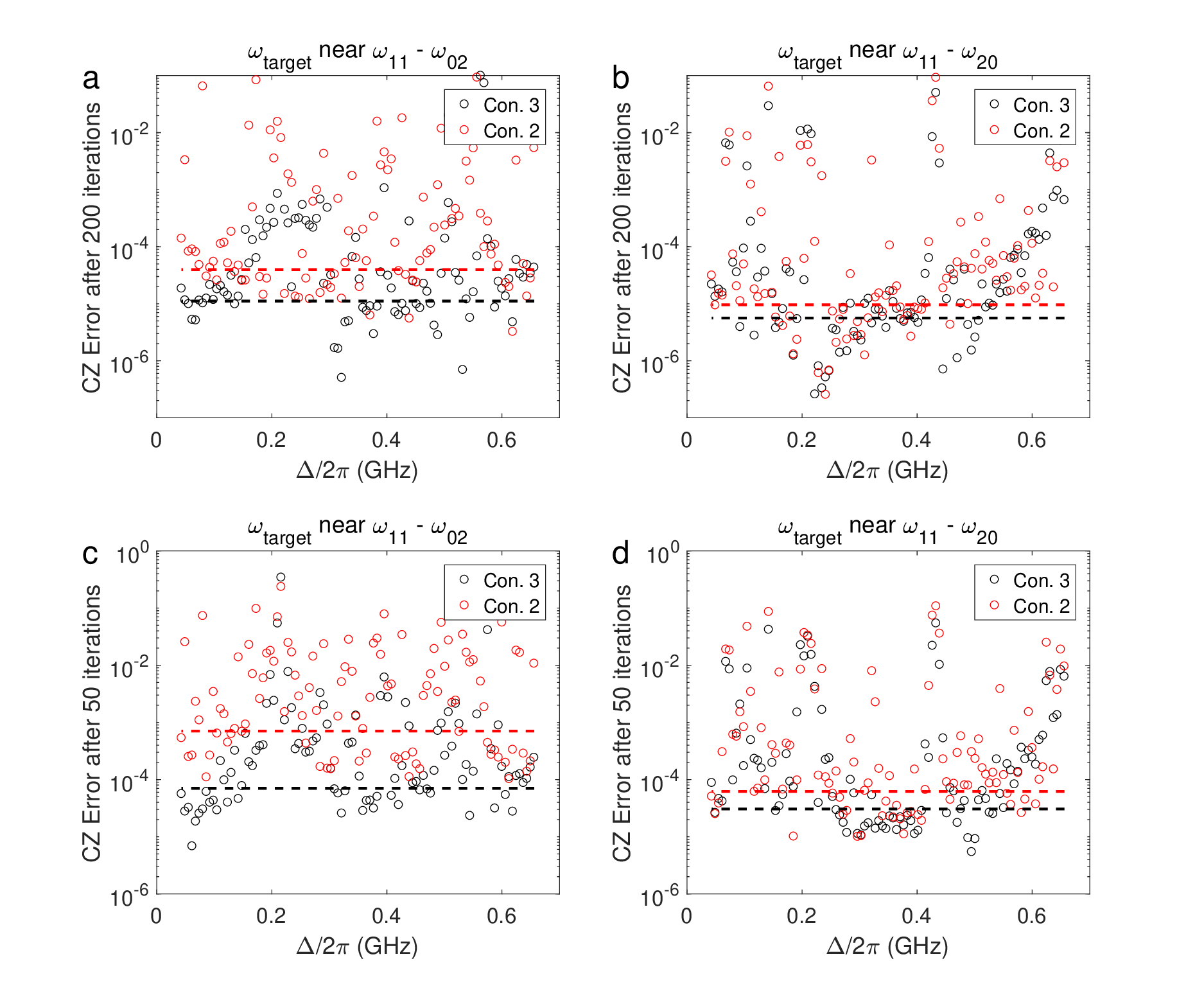}
    \caption{Results of Con. 3 and Con. 2 of CZ optimization.}
    \label{Figure_S3}
\end{figure*}

\begin{figure*}[hb]
    \centering
    \includegraphics[width=0.8\textwidth]{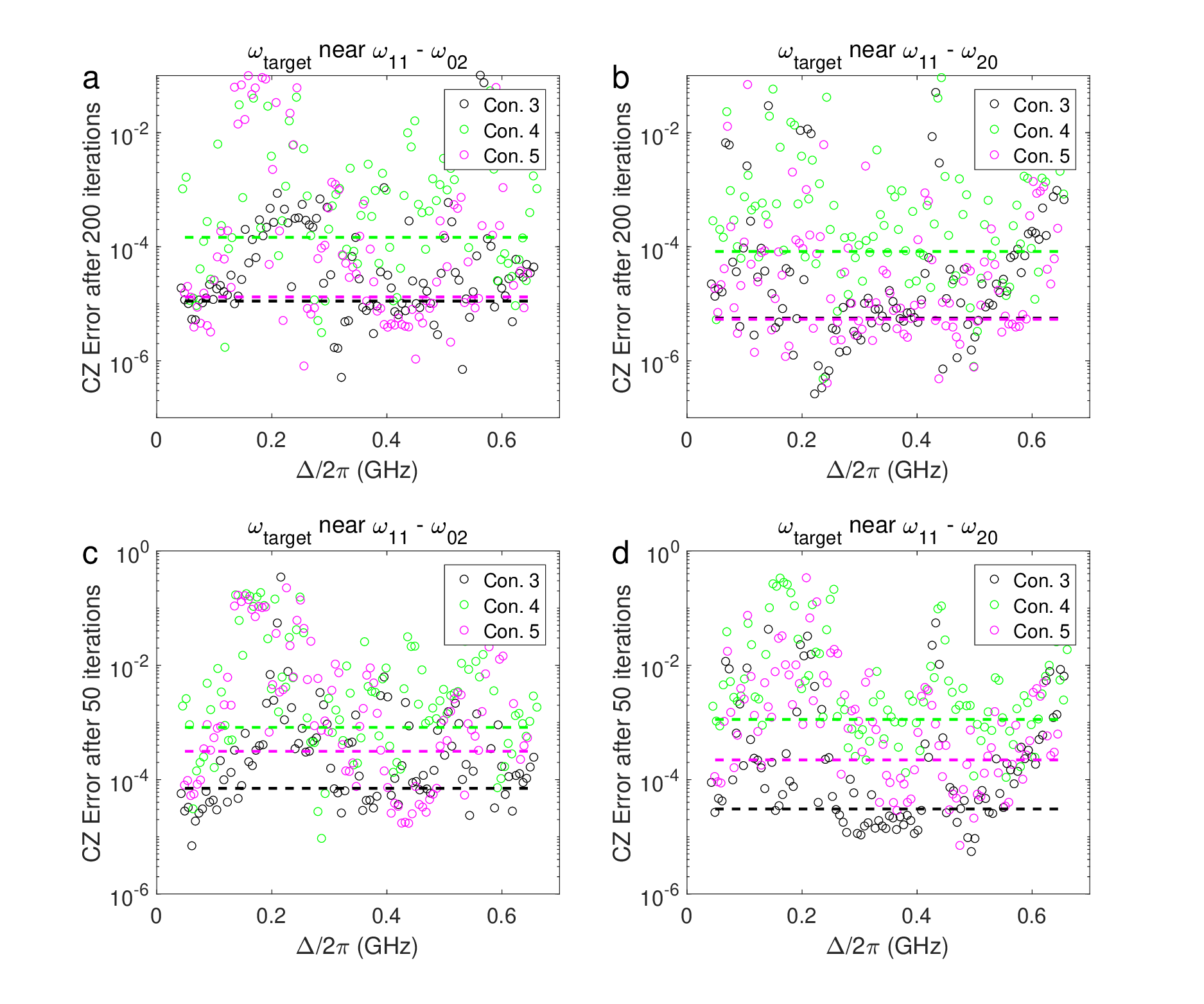}
    \caption{Results of Con. 3, Con. 4 and Con. 5 of CZ optimization.}
    \label{Figure_S4}
\end{figure*}

\begin{figure*}[hb]
    \centering
    \includegraphics[width=0.8\textwidth]{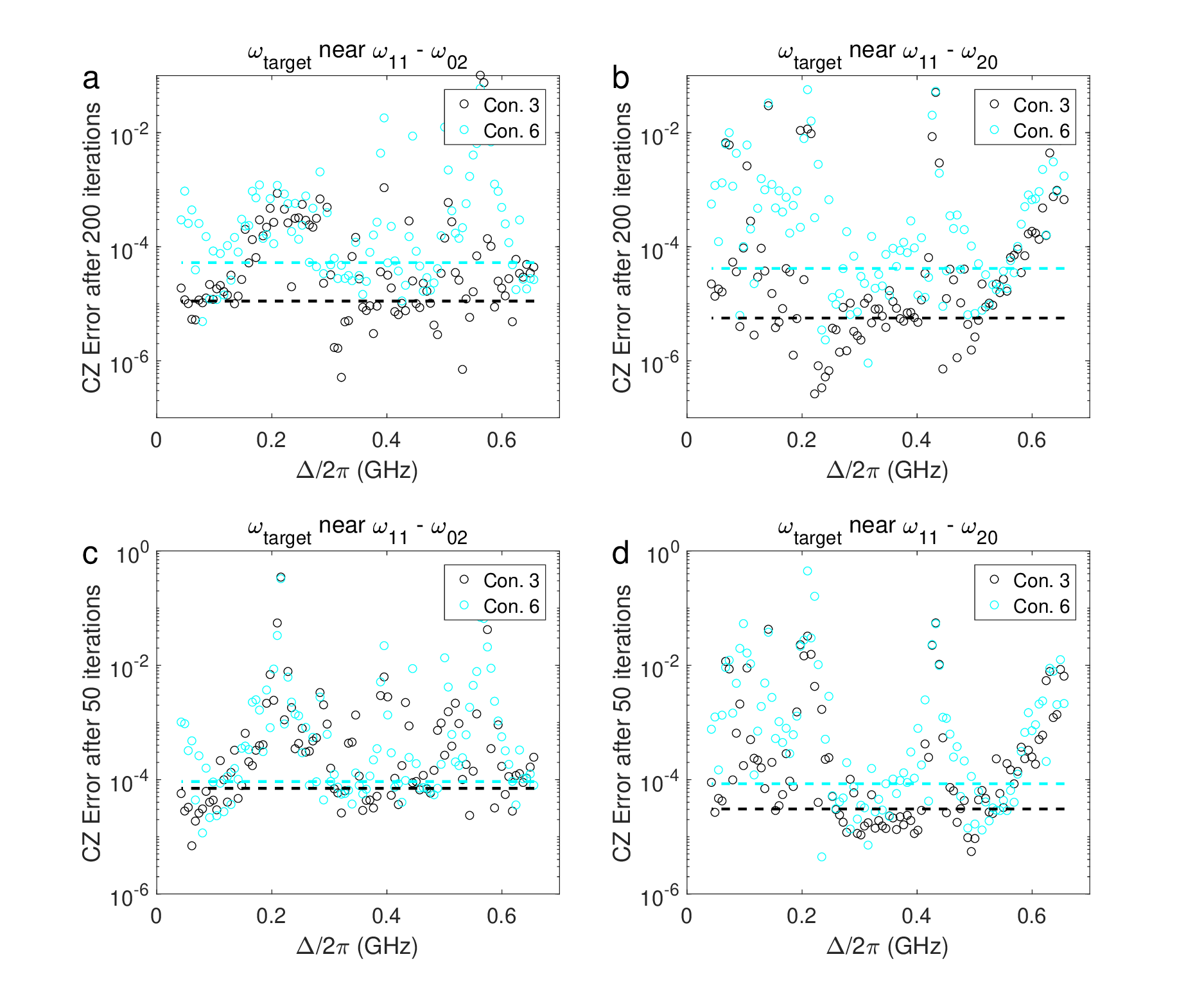}
    \caption{Results of Con. 3 and Con. 6 of CZ optimization.}
    \label{Figure_S5}
\end{figure*}

We designed three different couplers and compared the quality of different designs according to the CZ error achieved under different qubit detune frequency $\Delta= \omega_{10}-\omega_{01}$. We first set $g = 0$ for three types of the coupler, labeled as Con. 1, Con. 2, and Con. 3. We set different g for idle working points to find the effect of idle point setting, labeled as Con. 3, Con. 4, and Con. 5. Also, we test the influence of envelopes of the microwave by comparing two types of envelopes labeled as Con. 3, and Con. 6. Fig.\ref{Figure_S1} shows 6 condition for numerical simulation. Fig.\ref{Figure_S2} shows the different CZ error results between coupler design 1 and coupler design 3. Fig.\ref{Figure_S3} shows the different CZ error results between coupler design 2 and coupler design 3. Fig.\ref{Figure_S4} shows the results of different coupling strength setting for idle points. Fig.\ref{Figure_S5} shows the results of different initial envelope settings for idle points. 

\begin{figure*}[hb]
    \centering
    \includegraphics[width=0.9\textwidth]{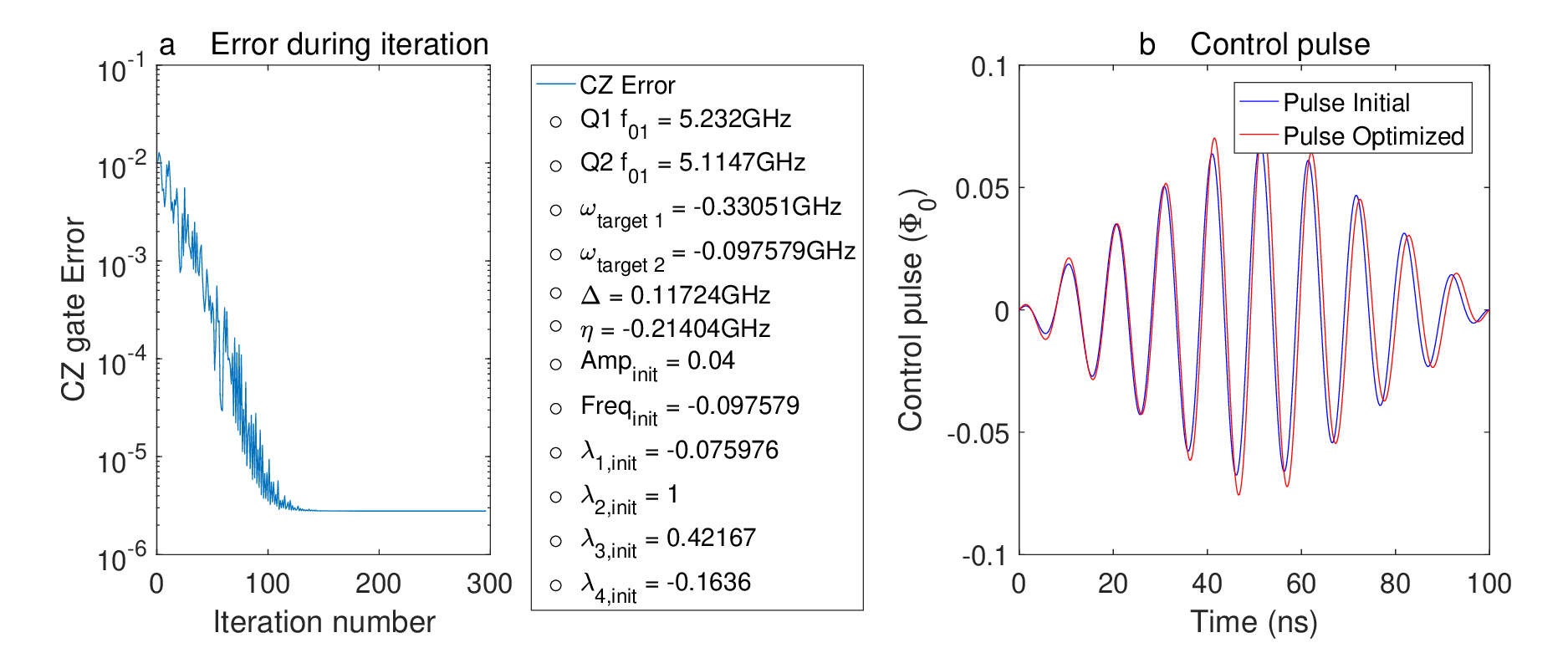}
    \caption{Numerical optimization at $\Delta/2\pi = 0.11724$ GHz in Fig. 2(d) in main text. (a) shows the qubits work point, initial value of CZ pulse and CZ gate error during iteration. (b) shows the control pulse before and after optimization.}
    \label{H1_1}
\end{figure*}

\begin{figure*}[hb]
    \centering
    \includegraphics[width=0.9\textwidth]{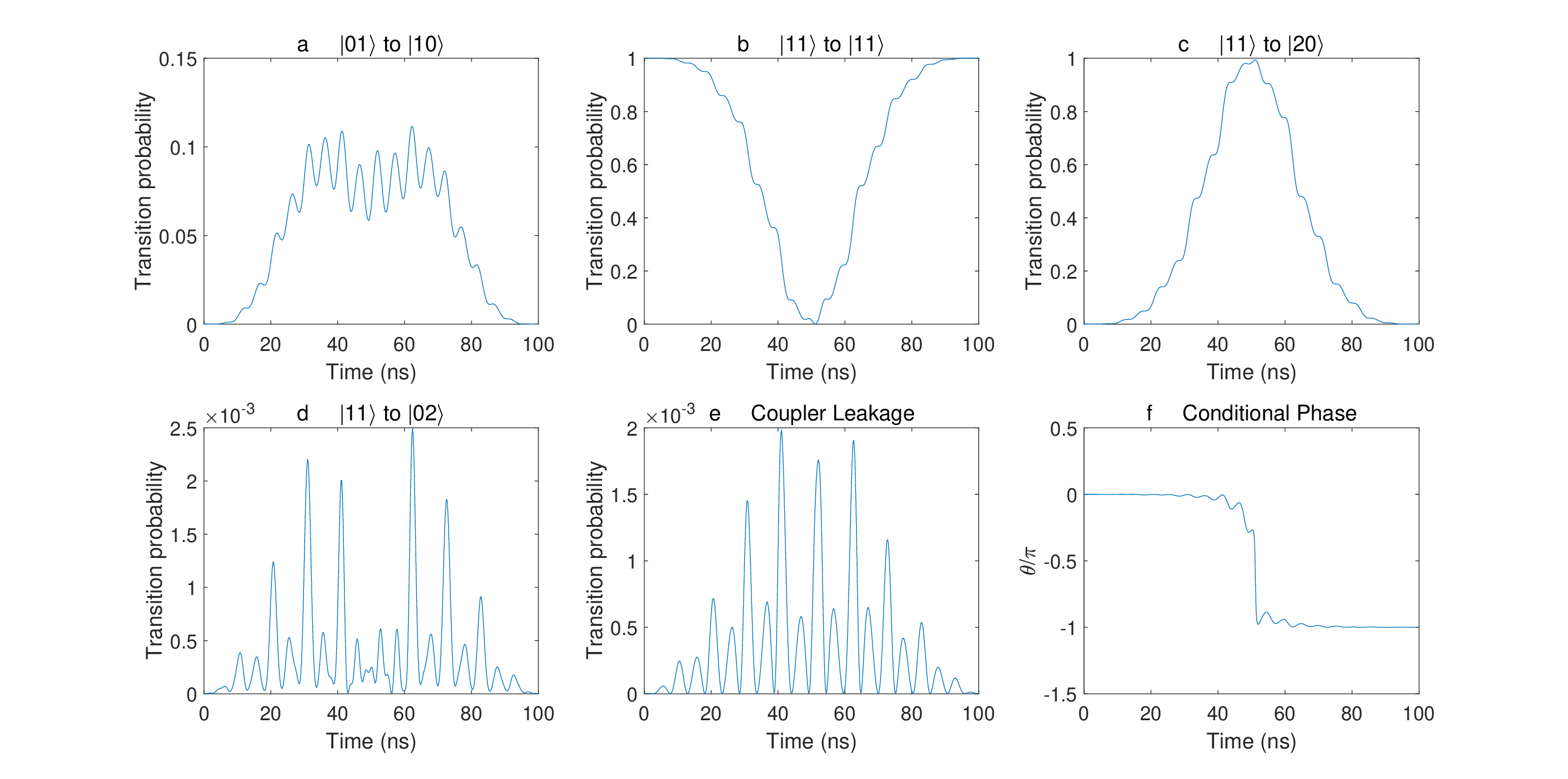}
    \caption{Time domain quantum evolution for optimized CZ pulse at $\Delta/2\pi = 0.11724$ GHz (data obtained by numerical simulation method). The quantum state in this figure is the energy eigenstate of the static Hamiltonian. (a)~(d) show the transtion probability between different quantum states. (e) shows the probability of leakage into the excited state of the coupler. (f) shows the conditional phase during the evolution.}
    \label{H1_2}
\end{figure*}

\begin{figure*}[hb]
    \centering
    \includegraphics[width=0.9\textwidth]{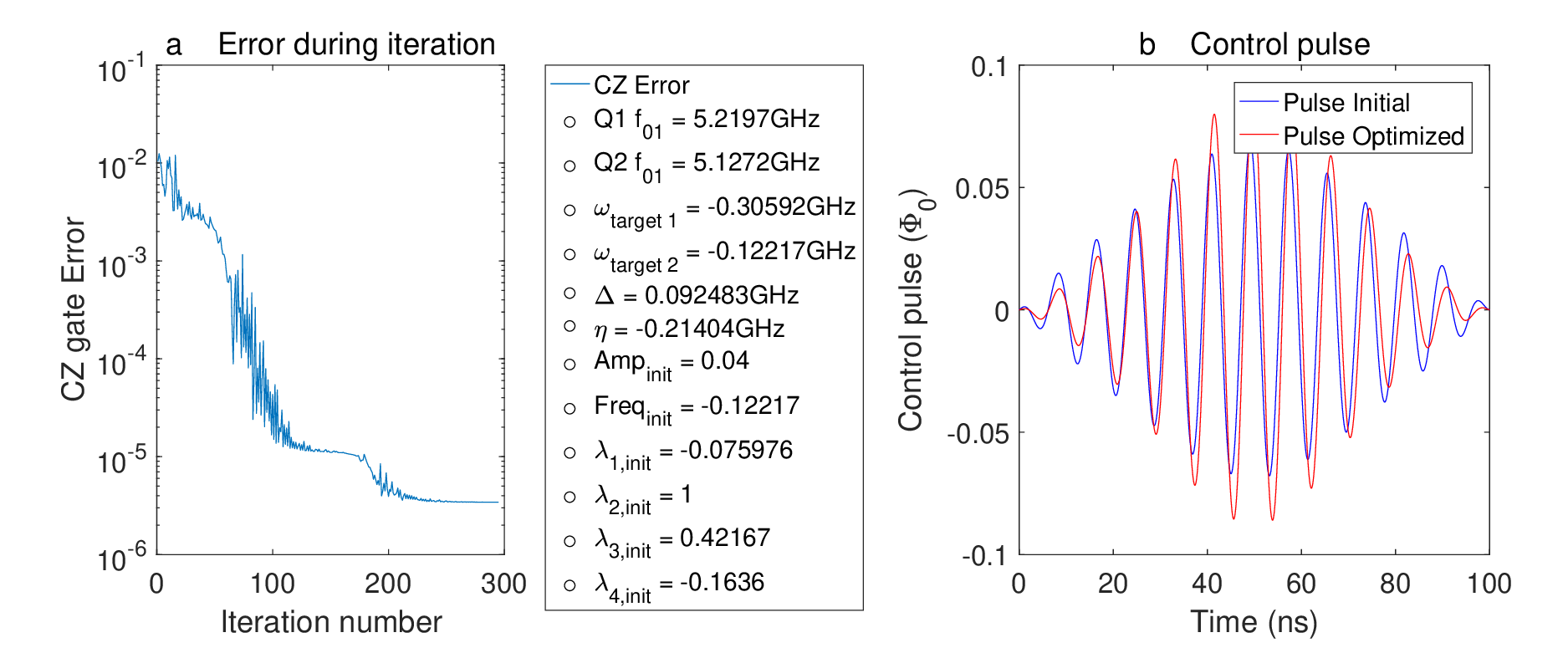}
    \caption{Numerical optimization at $\Delta/2\pi = 0.09248$ GHz in Fig. 2(d) in main text.}
    \label{H2_1}
\end{figure*}

\begin{figure*}[hb]
    \centering
    \includegraphics[width=0.9\textwidth]{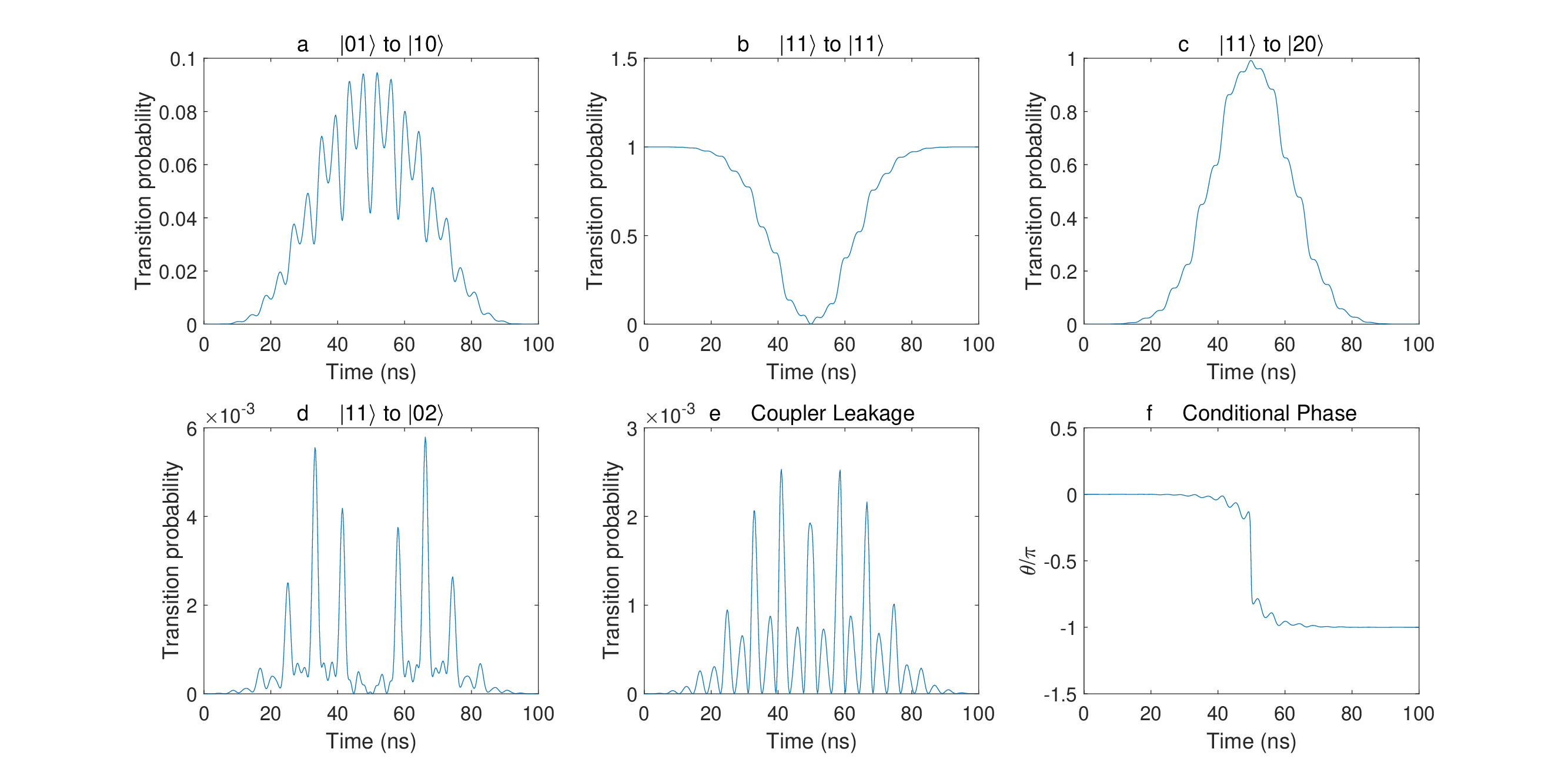}
    \caption{Time domain quantum evolution for optimized CZ pulse at $\Delta/2\pi = 0.09248$ GHz.}
    \label{H2_2}
\end{figure*}

\begin{figure*}[hb]
    \centering
    \includegraphics[width=0.9\textwidth]{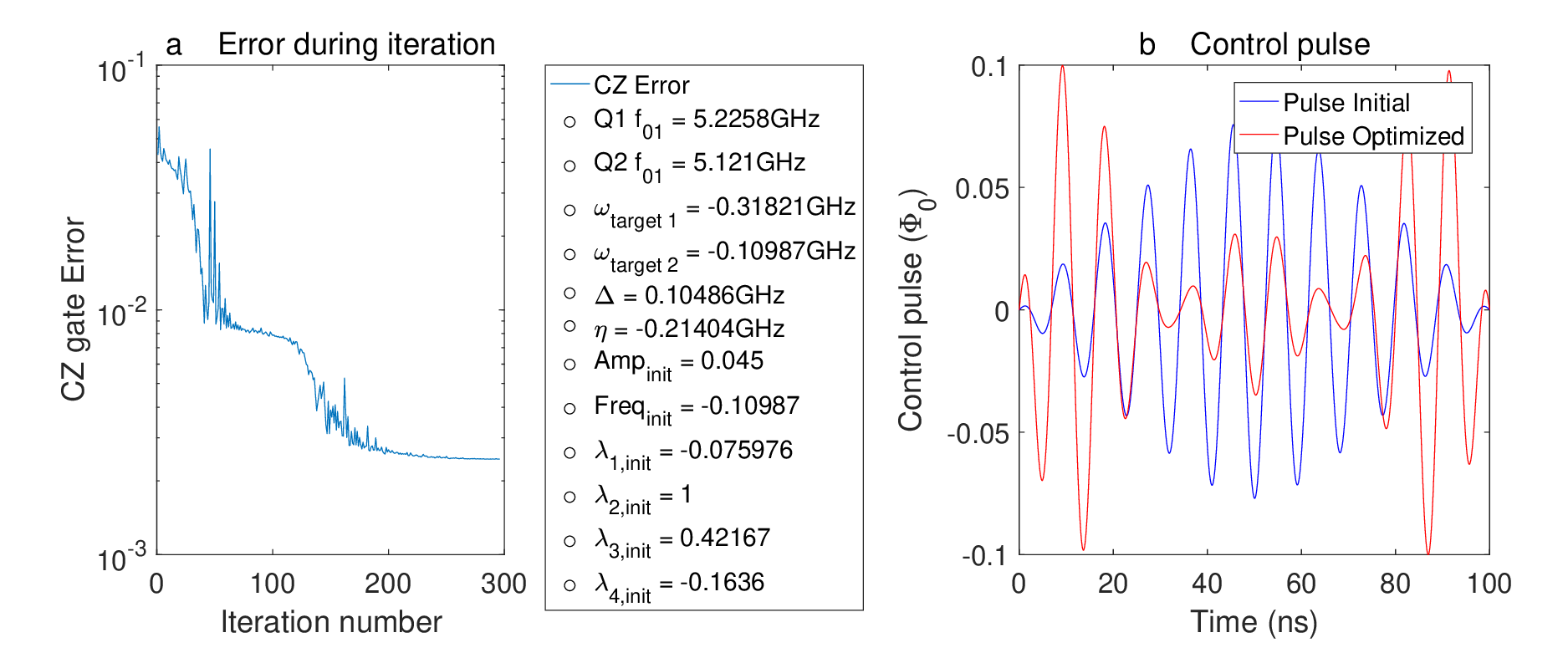}
    \caption{Numerical optimization at $\Delta/2\pi = 0.10486$ GHz in Fig. 2(d) in main text. For this $\Delta$, the interaction between $|01\rangle$ and $|10\rangle$ is activated.}
    \label{H3_1}
\end{figure*}

\begin{figure*}[hb]
    \centering
    \includegraphics[width=0.9\textwidth]{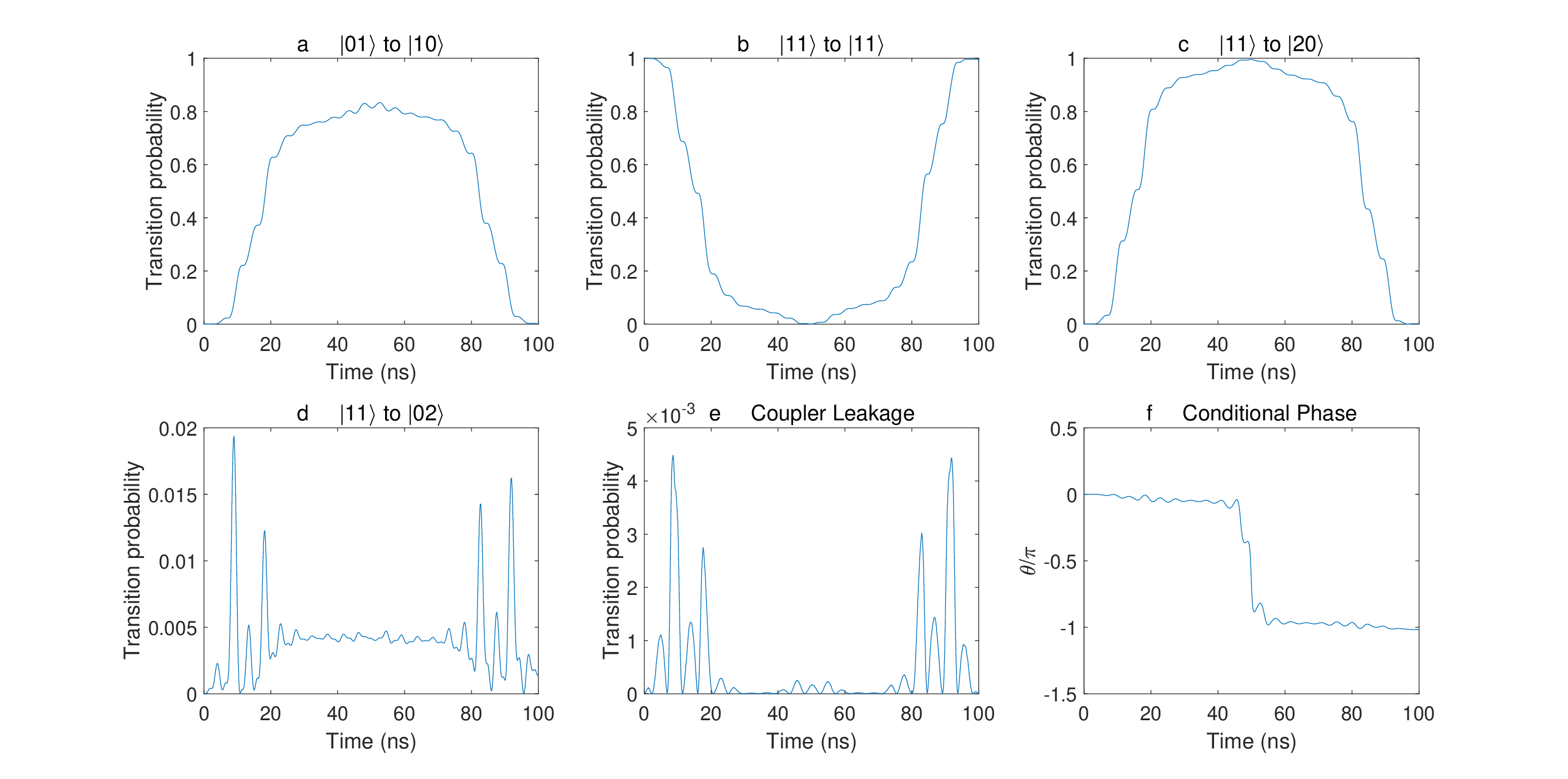}
    \caption{Time domain quantum evolution for optimized CZ pulse at $\Delta/2\pi = 0.10486$ GHz.}
    \label{H3_2}
\end{figure*}

\begin{figure*}[hb]
    \centering
    \includegraphics[width=0.9\textwidth]{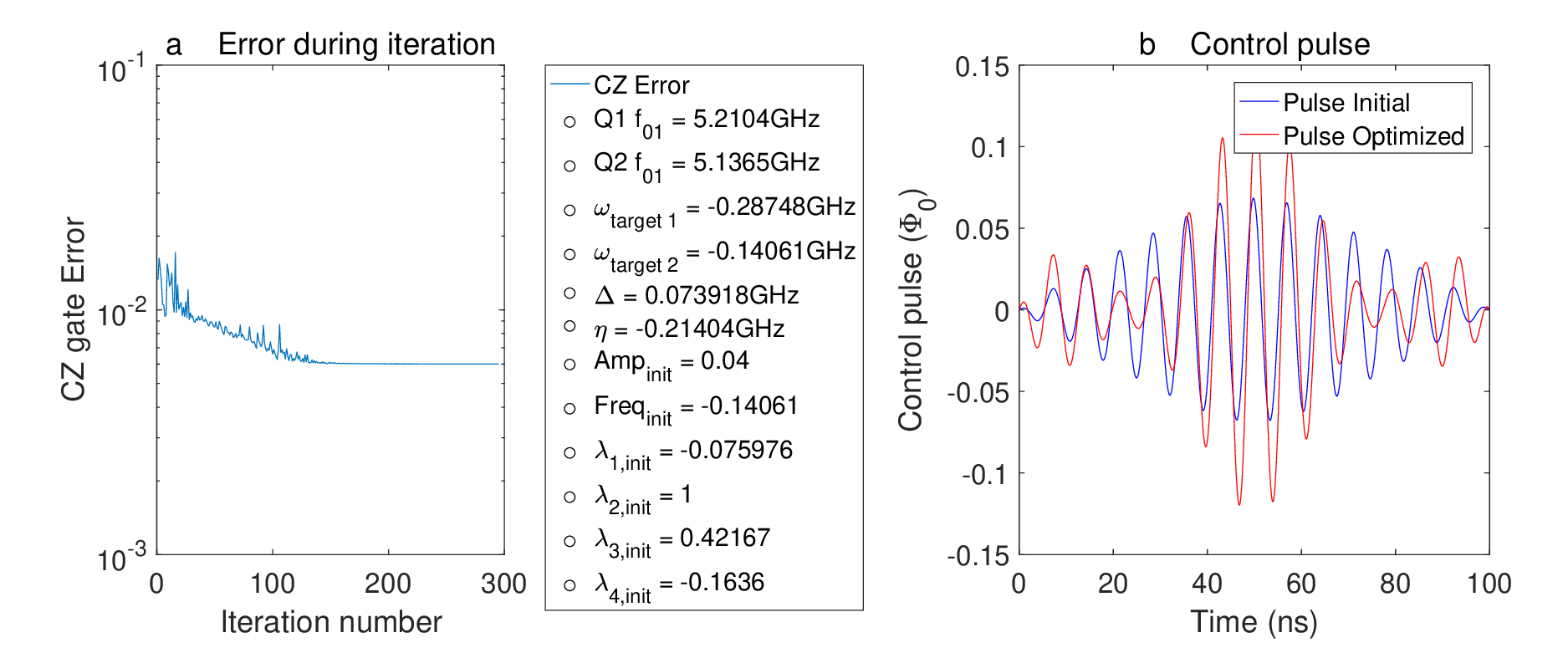}
    \caption{Numerical optimization at $\Delta/2\pi = 0.07392$ GHz in Fig. 2(d) in main text. For this $\Delta$, the interaction between $|11\rangle$ and $|02\rangle$ is activated.}
    \label{H4_1}
\end{figure*}

\begin{figure*}[hb]
    \centering
    \includegraphics[width=0.9\textwidth]{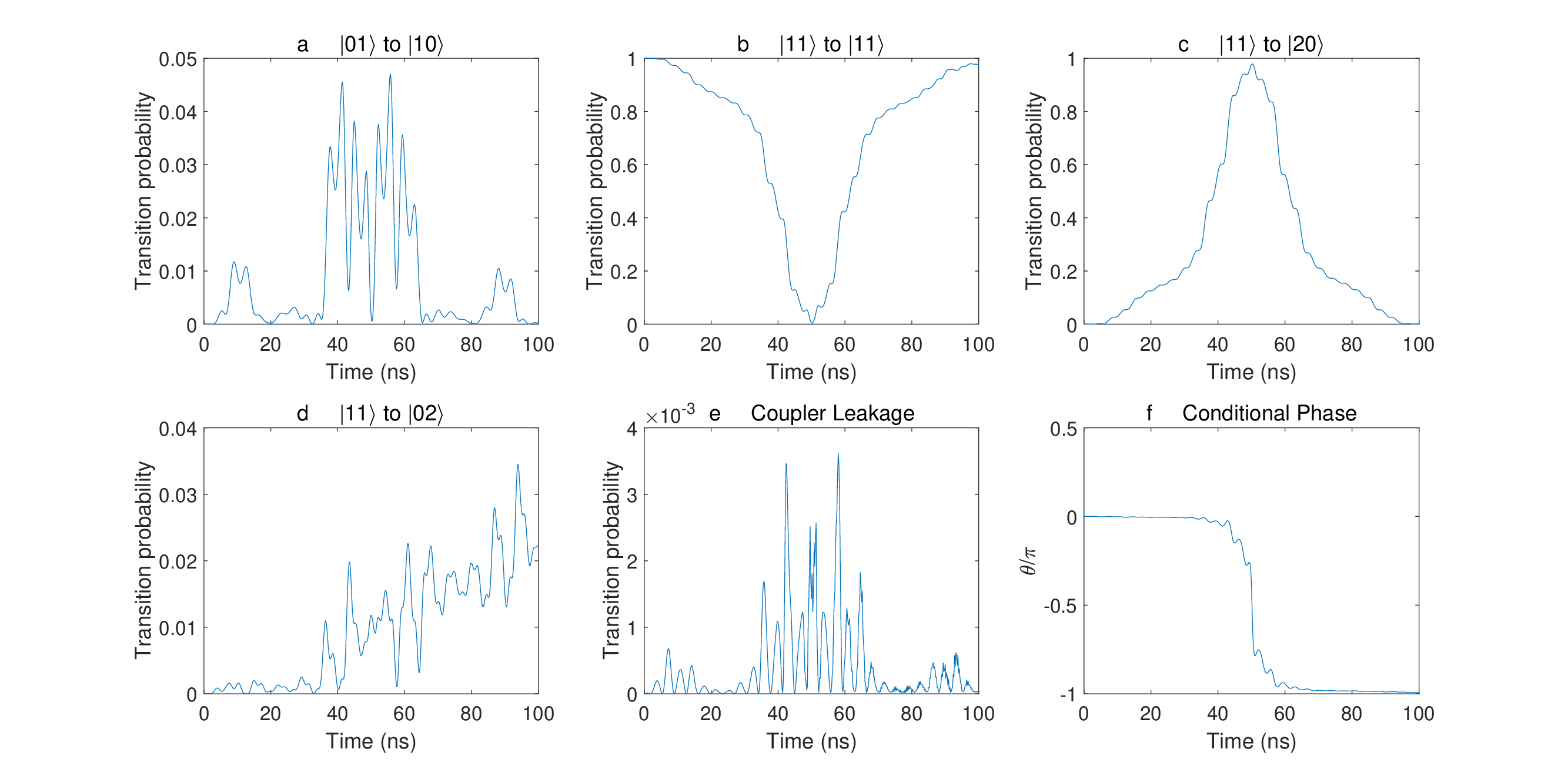}
    \caption{Time domain quantum evolution for optimized CZ pulse at $\Delta/2\pi = 0.07392$ GHz.}
    \label{H4_2}
\end{figure*}

For each $\Delta/2\pi$ in Fig. 4(d) in main text, we set the initial ratio of $\vec{\lambda}$ as $\lambda_1:\lambda_2:\lambda_3:\lambda_4 = -0.0760:1.0000:0.4222:-0.1636$, and chose static frequency difference between $|11\rangle$ and $|20\rangle$ as the initial microwave frequency. Also, the averaging microwave amplitude $\Bar{A}(t)$ was set to be 0.4, 0.045 or 0.05. Then we used the Nelder Mead algorithm to optimize the fidelity of the CZ gate. Fig.\ref{H1_1} and Fig.\ref{H1_2} show the numerical optimization and time domain evolution at $\Delta/2\pi = 0.11724$ GHz, where no unwanted interaction is activated and CZ gate with 99.99972\% fidelity is achieved. Fig.\ref{H2_1} and Fig.\ref{H2_2} show the numerical optimization and time domain evolution at $\Delta/2\pi = 0.09248$ GHz, where no unwanted interaction is activated and CZ gate with 99.99960\% fidelity is achieved. Fig.\ref{H3_1} and Fig.\ref{H3_2} show the numerical optimization and time domain evolution at $\Delta/2\pi = 0.10486$ GHz, where the interaction between $|01\rangle$ and $|01\rangle$ is activated and the optimized CZ fidelity is 99.75\%. Fig.\ref{H4_1} and Fig.\ref{H4_2} show the numerical optimization and time domain evolution at $\Delta/2\pi = 0.07391$ GHz, where the interaction between $|11\rangle$ and $|02\rangle$ activated and the optimized CZ fidelity is 99.39\%.

\section{Experiment of CZ gate}

The definition of cross entropy benchmarking (XEB) is as
below,

\begin{equation}
    \begin{aligned}
    \alpha = \frac{\overline{\sum_q{p_m(q)(Dp_s(q)-1)}}}{D\sum_q{\overline{p_s(q)^2}-1}}
    \end{aligned}
    \label{eq1}
\end{equation}
with $\alpha$ as the sequence fidelity, $D = 2^N$, $N$ is the number of qubits, $q$ is the sampled qubit state bitstrings (for single-qubit
XEB, $q$ is 0 or 1), $p_s(q)$ is the ideal probability of $q$, and $p_m(q)$ is the measured probability of $q$. The over lines in Eq. \ref{eq1} refer to the average of the random circuits in each cycle. 
Meanwhile, we use the speckle purity benchmarking (SPB) to measure the effect of decoherence error. Here,
\begin{equation}
    \begin{aligned}
    \sqrt{P} = \sqrt{Var(p_m)\frac{D^2(D+1)}{D-1}}.
    \end{aligned}
    \label{eq2}
\end{equation}
and $Var(P_m)$ the variance of the experimental probabilities extracted from the XEB experiment.

Due to the readout error, the start points (m = 0) of $\alpha$ are not at 1 in Fig.4(c) and (d). The setting of readout pulse in our experiment was different for $|0\rangle$, $|1\rangle$ measurement and for $|0\rangle$, $|1\rangle$, $|2\rangle$ measurement. For $|0\rangle$, $|1\rangle$ measurement, the matrixes of readout fidelity are labeled as below,
\begin{equation}
    \left(
    \begin{array}{cc}
        F_{00} & F_{01} \\
        F_{10} & F_{11}
    \end{array}
    \right)_{Q1}= 
    \left(
    \begin{array}{cc}
        0.9551 & 0.0449 \\
        0.1581 & 0.8419
    \end{array}
    \right)
    \label{eq3}
\end{equation}
and
\begin{equation}
    \left(
    \begin{array}{cc}
        F_{00} & F_{01} \\
        F_{10} & F_{11}
    \end{array}
    \right)_{Q2}= 
    \left(
    \begin{array}{cc}
        0.9475 & 0.0525 \\
        0.1517 & 0.8483
    \end{array}
    \right).
    \label{eq3}
\end{equation}
For $|0\rangle$, $|1\rangle$, $|2\rangle$ measurement, the matrixes of readout fidelity are labeled as below,
\begin{equation}
    \left(
    \begin{array}{ccc}
        F_{00} & F_{01} & F_{02}\\
        F_{10} & F_{11} & F_{12}\\
        F_{20} & F_{21} & F_{22}
    \end{array}
    \right)_{Q1}= 
    \left(
    \begin{array}{ccc}
        0.9050 & 0.0632 & 0.0318\\
        0.1633 & 0.7779 & 0.0588\\
        0.1383 & 0.2005 & 0.6612\\
    \end{array}
    \right)
    \label{eq3}
\end{equation}
and
\begin{equation}
    \left(
    \begin{array}{ccc}
        F_{00} & F_{01} & F_{02}\\
        F_{10} & F_{11} & F_{12}\\
        F_{20} & F_{21} & F_{22}
    \end{array}
    \right)_{Q2}= 
    \left(
    \begin{array}{ccc}
        0.8767 & 0.0534 & 0.0699\\
        0.1887 & 0.7399 & 0.0714\\
        0.1645 & 0.1784 & 0.6571\\
    \end{array}
    \right).
    \label{eq3}
\end{equation}
Here $F_{ij}$ represents the probability of measuring $|j\rangle$ when we prepared the initial state of $|i\rangle$. In experiment, we used the product of each matrix of single-qubit readout fidelity as the matrix of multi-qubit readout fidelity. In Fig.4(c) and (d), the $\alpha$ and $\sqrt{P}$ are calculated by the raw probability of $|0\rangle$, $|1\rangle$ measurement, and the $leak$ is the probability corrected according to the matrix of multi-qubit readout fidelity of $|0\rangle$, $|1\rangle$, $|2\rangle$ measurement. Here, $leak = P_{|02\rangle}+P_{|12\rangle}+P_{|20\rangle}+P_{|21\rangle}+P_{|22\rangle}$. 

